\ifdefined\icraversion
  \documentclass[letterpaper,10pt,conference]{ieeeconf}
\else
  \documentclass[11pt]{article}
\fi
\ifdefined\icraversion
\fi

\ifdefined\icraversion
  \usepackage{cite}
\else
  \usepackage[letterpaper,margin=1.05in]{geometry}
  \usepackage{newtxtext}
  \usepackage{newtxmath}
  \usepackage[numbers]{natbib}
  \usepackage{multicol}
  \usepackage{titlesec}
  \usepackage{needspace}
\fi
\usepackage{microtype}
\usepackage{url}

\ifdefined\icraversion\else

  \usepackage[labelfont=bf]{caption}
  \usepackage[bookmarks=true,
    pdftitle={Aero Hand Open: A Simulation-Ready Tendon-Driven Hand for Dexterous Manipulation Learning},
    pdfauthor={Nan Wang, Mohit Yadav, Jonathan Wulff, Aidan Rosenbaum, Kezhou Chen, Yuvan Sharma, Xu Dong, Yiwei Tao},
    pdfsubject={Robotics},
    pdfkeywords={Dexterous manipulation; Tendon-driven hand; Sim-to-real; Reinforcement learning}]{hyperref}
\fi
\usepackage{amsmath}
\ifdefined\icraversion\else
  
\fi
\usepackage{amssymb}
\usepackage{graphicx}
\ifdefined\icraversion
  \usepackage[caption=false,font=footnotesize]{subfig}
\else
  \usepackage{subfig}
\fi
\usepackage{xcolor}

\newcommand{\assign}[2]{\ifx\showassign\undefined\else
  {\color{#1}\bfseries\sffamily [\,#2\,]}\fi}

\usepackage{comment}
\ifdefined\confversion
  
  \excludecomment{reportonly}
  \includecomment{confonly}
  \newcommand{\taxwidth}{0.84\linewidth}
  \newcommand{\reportonlytext}[1]{}
  \newcommand{\confonlytext}[1]{#1}
\else
  
  \includecomment{reportonly}
  \excludecomment{confonly}
  \newcommand{\taxwidth}{0.84\textwidth}
  \newcommand{\reportonlytext}[1]{#1}
  \newcommand{\confonlytext}[1]{}
\fi

\definecolor{mohitgreen}{rgb}{0,0.45,0.10}
\newcommand{\revised}[1]{\ifx\showrevised\undefined#1\else{\color{blue}#1}\fi}
\newcommand{\mohit}[1]{\ifx\showrevised\undefined#1\else{\color{mohitgreen}#1}\fi}
\newcommand{\rtwo}[1]{\ifx\showrevised\undefined#1\else{\color{red}#1}\fi}
\newenvironment{rtwoblock}%
  {\ifx\showrevised\undefined\else\color{red}\fi}{}
\newenvironment{mohitblock}%
  {\ifx\showrevised\undefined\else\color{mohitgreen}\fi}{}
\usepackage{booktabs}

\definecolor{aeroblue}{rgb}{0.18,0.36,0.62}
\ifdefined\icraversion\else
  \titleformat{\section}{\normalfont\Large\bfseries\sffamily\color{aeroblue}}
    {\thesection}{0.8em}{}
  \titleformat{\subsection}{\normalfont\large\bfseries\sffamily}
    {\thesubsection}{0.7em}{}
  \titleformat{\subsubsection}{\normalfont\normalsize\bfseries\sffamily}
    {\thesubsubsection}{0.6em}{}
  \titlespacing*{\section}{0pt}{1.6ex plus 0.6ex minus 0.2ex}{0.9ex plus 0.2ex}
  \titlespacing*{\subsection}{0pt}{1.3ex plus 0.5ex minus 0.2ex}{0.7ex plus 0.2ex}
\fi

\newcommand{\grouphead}[1]{\par\medskip\noindent{\sffamily\bfseries #1}\par\nobreak\smallskip\noindent\ignorespaces}

\ifdefined\icraversion
  \newenvironment{widefigure}[1][t]{\begin{figure*}[#1]}{\end{figure*}}
  \newenvironment{widetable}[1][t]{\begin{table*}[#1]}{\end{table*}}
\else
  \newenvironment{widefigure}[1][htbp]{\begin{figure}[!htbp]}{\end{figure}}
  \newenvironment{widetable}[1][htbp]{\begin{table}[!htbp]}{\end{table}}
\fi

\ifdefined\anonymousversion
  \newcommand{\HandName}{The proposed hand}
  \newcommand{\handname}{the proposed hand}
\else
  \newcommand{\HandName}{Aero Hand Open}
  \newcommand{\handname}{Aero Hand Open}
\fi

\graphicspath{{figures/}{logos/}}

\begin{document}

\ifdefined\icraversion
  \title{A Simulation-Ready Tendon-Driven Hand for\\Dexterous Manipulation Learning}
  \author{}
\else
  \title{%
    \vspace{-5.0em}%
    \includegraphics[height=8mm]{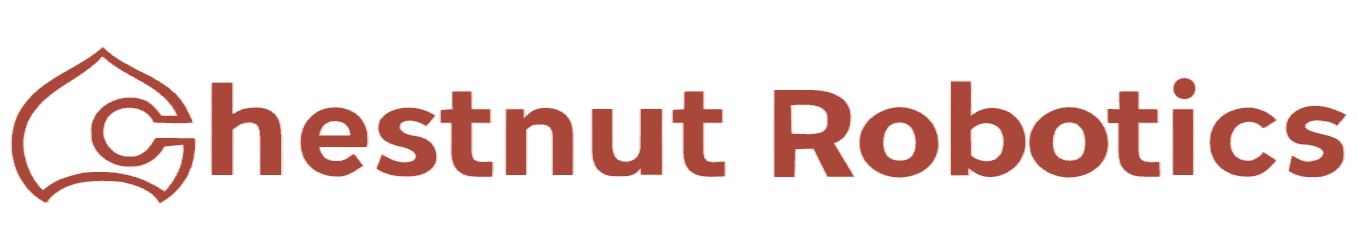}\\[0.9em]
    \bfseries Aero Hand Open: A Simulation-Ready Tendon-Driven Hand\\for Dexterous Manipulation Learning%
    \vspace{-0.2em}}
  \author{%
    \rtwo{Nan Wang\textsuperscript{1} \quad Mohit Yadav\textsuperscript{1} \quad Jonathan ``Jono'' Wulff\textsuperscript{1} \quad Aidan Rosenbaum\textsuperscript{1}
   \quad Kezhou Chen\textsuperscript{1}\\[0.25em]
    Yuvan Sharma\textsuperscript{1,2,\dag} \quad Xu ``Joe'' Dong\textsuperscript{1,*} \quad Yiwei ``Evan'' Tao\textsuperscript{1,*}\\[0.5em]
    \normalsize\textsuperscript{1}Chestnut Robotics\textsuperscript{\ddag} \qquad
    \textsuperscript{2}California Institute of Technology}\\[0.55em]
    \normalsize\url{https://chestnut-robotics.github.io/aero-hand-open/}}
  \date{}
\fi

\newcommand{\taxonomycaption}{\HandName{} performing the $33$ grasp types of the GRASP taxonomy \cite{feix2016grasp}, numbered in taxonomy order. Frame colour gives the category: power (red), intermediate (orange) and precision (blue). All grasps use the seven motors of the hand with no change of hardware between grasps.}
\ifdefined\icraversion
  \makeatletter
  \let\aho@titleonly\@maketitle
  \renewcommand{\@maketitle}{%
    \aho@titleonly
    \vskip 0.7em%
    \centerline{\includegraphics[width=\textwidth]{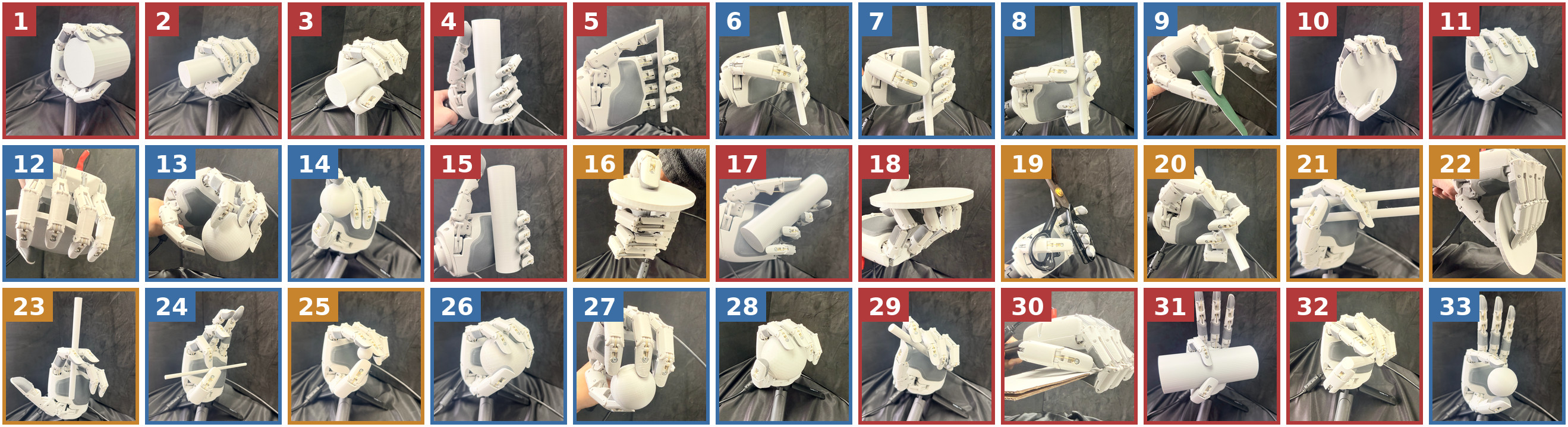}}%
    \def\@captype{figure}%
    \@makecaption{\fnum@figure}{\taxonomycaption}%
    \vskip 0.3em%
  }
  \refstepcounter{figure}
  \makeatother
\fi

\maketitle

\ifdefined\icraversion\label{fig:taxonomy}\fi

\ifdefined\icraversion\else
  \begingroup
  \renewcommand{\thefootnote}{}%
  \footnotetext{\rtwo{\textsuperscript{*}Corresponding authors. Software: Xu ``Joe'' Dong, \texttt{joedong@chestnut.bot}. Mechanical design: Yiwei ``Evan'' Tao, \texttt{evantao@chestnut.bot}.}}
  \footnotetext{\rtwo{\textsuperscript{\dag}Yuvan Sharma was an intern at Chestnut Robotics during this work, and is now at the California Institute of Technology.}}
  \footnotetext{\rtwo{\textsuperscript{\ddag}Chestnut Robotics has rebranded from TetherIA; of the project links, only the documentation site retains the earlier name.}}
  \endgroup
\fi

\begin{abstract}
\revised{Tendon-driven hands are anthropomorphic, and moving the actuators off the joints is what makes a hand of this capability affordable to build.}
\rtwo{Two effects produce that saving.
Routing force through a cable removes the requirement that a motor fit inside the joint it drives, so smaller and cheaper motors suffice, and one motor can drive several joints through a single cable, so fewer motors are needed.}
\revised{They are also harder to learn on than a direct-drive hand.}
\rtwo{The underactuated transmission that produces the saving is itself difficult to represent in a simulator, and the joints one cable drives are not independently commandable.
We present \textbf{\handname}, a tendon-driven anthropomorphic hand that is released simulation-ready.
Three things ship with it.
A simulation model reproduces the cable transmission itself.
An identified actuation map connects that model to the motor commands in both directions, including the three-way coupling of the thumb.
A reinforcement learning package trains policies for the hand.
Together they let a policy be trained entirely in simulation and run on the hand with no fine-tuning and no state estimation.
We release the mechanical design, the simulation model, the identified mapping, the training environment and the deployment stack.}
\end{abstract}

\ifdefined\icraversion\else
\begin{widefigure}[!t]
    \centering
    \includegraphics[width=\taxwidth]{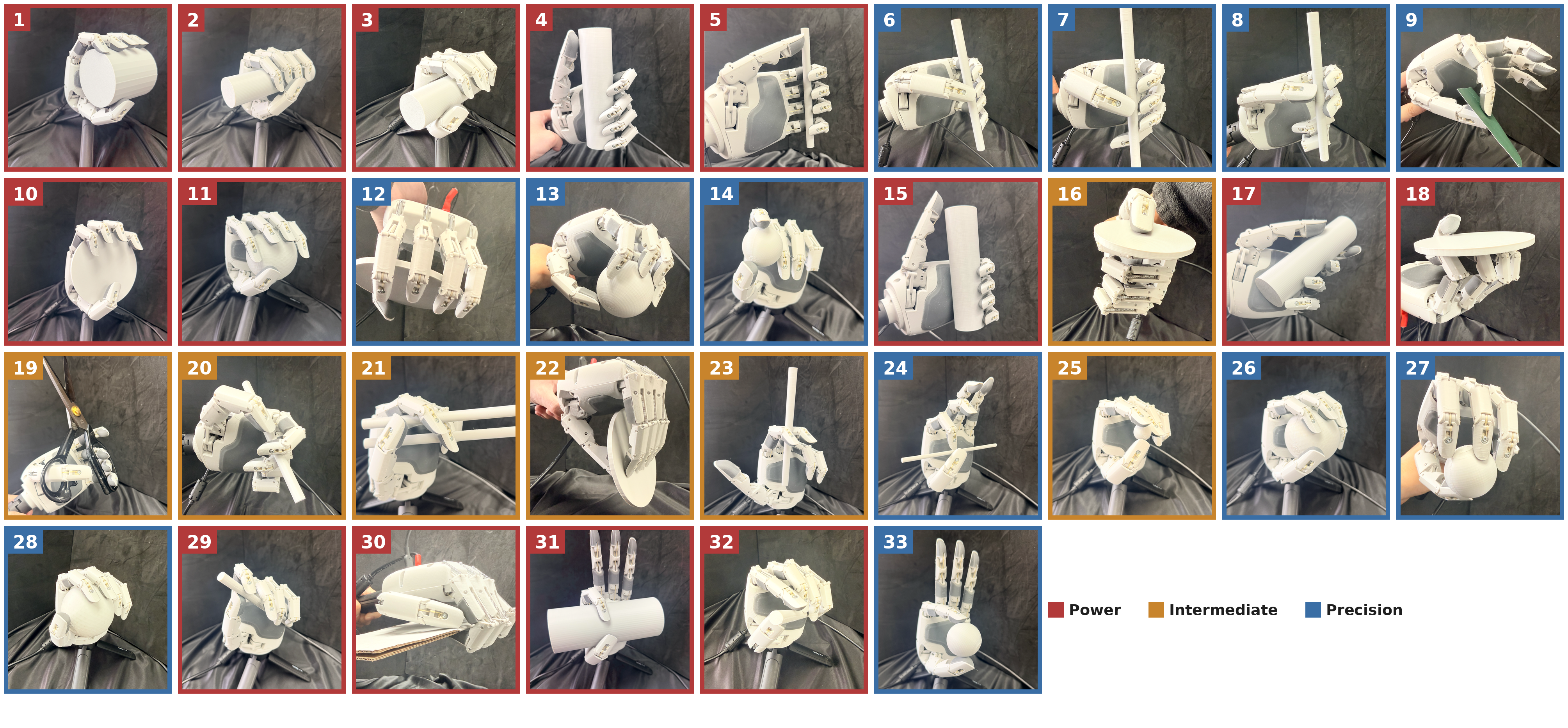}
    \caption{\taxonomycaption}
    \label{fig:taxonomy}
\end{widefigure}
\fi

\section{Introduction}
\rtwo{Dexterous manipulation has been a long-standing challenge in robotics, in part because contacts make and break as the object moves in the hand, so the dynamics switch between modes rather than evolving smoothly \cite{mordatch2012cio, posa2014direct, hogan2020reactive, wang2025hybridplanning}.
This paper trains policies in simulation, and what that route demands of the hardware is a simulation model faithful enough to train in.}
\reportonlytext{Robotic arms with grippers are widely deployed in industrial assembly \cite{industrial}, robotic surgery \cite{surgerygripper}, space exploration \cite{spacegripper}, micro-manipulation \cite{micro}, and agriculture \cite{blackberry}, and two-finger and three-finger grippers suffice for many of those tasks.}\confonlytext{Two-finger and three-finger grippers suffice for many industrial tasks.}
Precise or adaptive manipulation still calls for multi-finger hands.
Grasping irregularly shaped objects and rotating an object in the hand both require the cooperation of several fingers \cite{qihaozhi-arllegro, piano}.
Human tools are also designed for a hand that can oppose the thumb against multiple fingers for stable torque balance \cite{realdex}.
Hands of that kind, however, have long been either expensive or hard to learn on.

Robotic hands employ one of three actuation principles.
Tendon-driven transmission most closely mimics the human hand, placing the actuators in the forearm and transmitting force through cables.
On-joint servo actuation, as in the LEAP Hand \cite{shaw2023leaphand}, trades anthropomorphism for a direct and fully observable joint interface.
Pneumatically actuated hands such as the RBO Hand~2 \cite{deimel2016rbohand} obtain compliance and robustness from soft continuum structures.
The Shadow Dexterous Hand \cite{shadowdexterous} reproduces human hand kinematics with 24 movements and rich sensing, but at a prohibitive cost.
The ILDA \cite{ILDA} and ByteDexter \cite{bytedexter} hands achieve high dexterity through linkage transmissions, and the ISyHand \cite{isyhand} does so through an articulated palm.
The InMoov \cite{inmoov-improved} and Amazing Hand \cite{amazinghand} show that useful hands can be built for a few hundred dollars.
Closest to this work are the tendon-driven hands built for learning.
The Faive Hand \cite{toshimitsu2023faive} and ORCA \cite{christoph2025orca} transfer reinforcement learning policies from simulation to hardware without fine-tuning, ORCA with tactile sensing and an actuated wrist; RUKA \cite{zorin2025ruka} learns joint-to-actuator and fingertip-to-actuator models of an underactuated tendon hand from motion-capture data, and RUKA-v2 \cite{liu2026rukav2} extends that design with a wrist and finger abduction; and Yuryev and Hughes \cite{yuryev2026tendonforce} learn a tendon-force model to narrow the sim-to-real gap of a tendon-driven finger.
These hands drive their joints from eleven or more actuators.
The hand presented here drives sixteen joints from seven, senses only at the motors and, rather than learning the map from joints to actuators, represents the cable routing itself in the simulator and identifies the map from cable length to motor command channel by channel.
Table~\ref{tbl:handcompare} compares these platforms.

\begin{table}[!t]
    \centering
    \ifdefined\icraversion\footnotesize\setlength{\tabcolsep}{3pt}\fi
    \begin{tabular}{|l|c|c|c|c|c|} \hline
         Hand & Drive & Fingers & DoFs & Mass [g] & Price [\$] \\ \hline
         InMoov & Tendon & 5 & 5 & 750 & $\sim 200$ \\
         Allegro V5 Plus & Servo & 4 & 16 & 1,024 & 17,000 \\
         ISyHand & Servo & 4 & 18 & 620 & $\sim 1,300$ \\
         ILDA Hand & Linkage & 5 & 15 & 1,100 & NA \\
         LEAP Hand & Servo & 4 & 16 & 600 & 2,000 \\
         Shadow Hand & Tendon & 5 & 20 & 4,300 & $>100,000$ \\
         ByteDexter Hand & Linkage & 5 & 15 & 1,300 & NA \\
         Amazing Hand & Servo & 4 & 8 & 400 & $\sim 231$ \\
         Faive Hand & Tendon & 5 & 11 & 1,100 & NA \\
         ORCA & Tendon & 5 & 17 & 1,200 & $<$2,000 CHF \\
         RUKA & Tendon & 5 & 11 & NA & $<$1,300 \\
         \HandName & Tendon & 5 & 7 & 374 & 314 \\ \hline
    \end{tabular}
    \caption{Comparison of existing dexterous hands. DoFs denotes the number of actuated degrees of freedom, and NA marks a figure the manufacturer does not publish.}
    \label{tbl:handcompare}
\end{table}

We present \handname, a five-finger tendon-driven hand with sixteen revolute joints, seven motors, a mass of $374$\,g and a bill of materials of \$314.
Table~\ref{tbl:handcompare} places it among existing platforms.
Its mechanical capability is broad.
The hand covers all $33$ grasp types of the GRASP taxonomy \cite{feix2016grasp} with a single hardware configuration (Fig.~\ref{fig:taxonomy}).
Its structure is entirely 3D printed, and the design files are released along with the firmware and the control stack.

\rtwo{The hand is designed for low cost, and that decision propagates.
Seven motors are what the budget and the volume behind the palm allow, so sixteen joints are driven from seven actuators and the hand is underactuated by construction rather than by choice.
That is what has to be paid for in the simulator, and it is the gap this paper addresses.
The joints a cable drives are not independently commandable.
The only proprioception is the motor encoder.}
The cables also introduce couplings, friction and slack that a model built from independent joint-position actuators cannot represent.
The same observation motivated actuator-level modelling in legged locomotion \cite{hwangbo2019learning}, where identifying the transmission was the decisive step for zero-shot transfer.
Our contributions are:
\begin{enumerate}
    \item \rtwo{\textbf{An open-sourced, capable and repairable hand.} \rtwo{\HandName{} performs all $33$ grasp types of the GRASP taxonomy at $374$\,g.} Its mechanical design, firmware and control stack are released in full, so a worn or broken part is reprinted rather than reordered.}
    \item \rtwo{\textbf{A tendon-level simulation model.} We model the complete cable transmission of \handname{} in MuJoCo \cite{todorov2012mujoco}, with every actuation cable, coupling cable and return spring routed over wrapping geometries taken from the CAD model.}
    \item \rtwo{\textbf{An identified actuation map, validated in both directions.} We express the sim-to-real interface as a chain of linear maps that carries the thumb's three-way coupling, and validate it against the hardware channel by channel, kinematically and dynamically. Together with domain randomisation, this is what lets the policy transfer with no fine-tuning.}
    \item \rtwo{\textbf{A reinforcement learning (RL) package.} We release the training environment, the reward and randomisation settings, and the deployment node, with observation and action spaces restricted to hardware-available signals. We demonstrate the pipeline on in-hand cube rotation.}
\end{enumerate}

\section{Hand Design}
\label{sec:platform}

\begin{widefigure}[!t]
    \centering
    \subfloat[Exploded view\label{fig:exploded}]{%
        \parbox[c][2.62in][c]{0.43\textwidth}{\centering
        \includegraphics[height=2.50in]{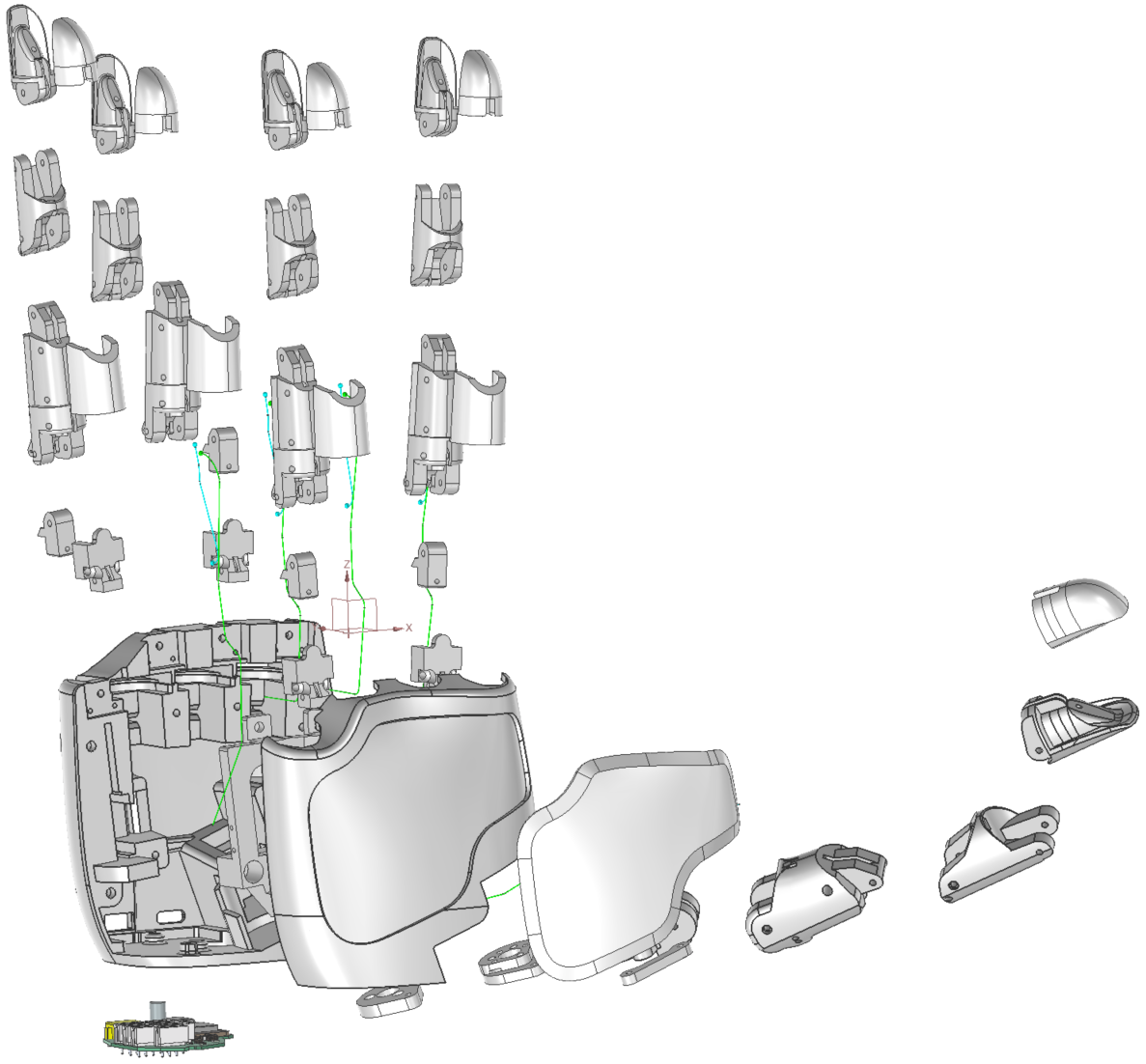}}}\hfil
    \subfloat[Finger joints\label{fig:finger}]{%
        \parbox[c][2.62in][c]{0.12\textwidth}{\centering
        \includegraphics[height=2.55in]{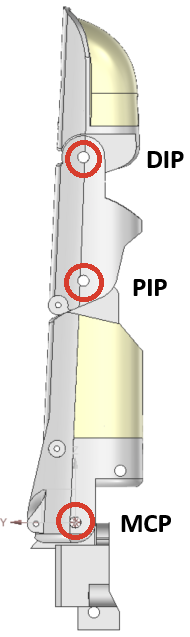}}}\hfil
    \subfloat[Thumb joints\label{fig:thumb}]{%
        \parbox[c][2.62in][c]{0.42\textwidth}{\centering
        \includegraphics[width=0.41\textwidth]{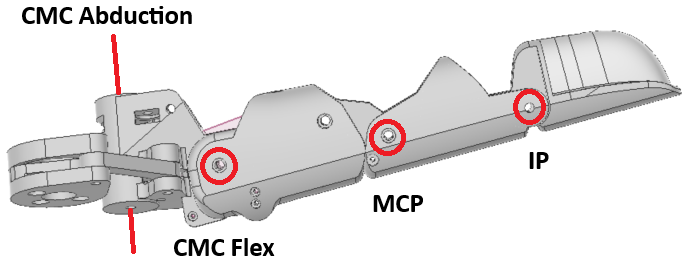}}}
    \caption{Mechanical design of \handname.
    (a)~Exploded view: the four fingers and the opposable thumb assemble onto a two-part printed palm that houses the seven servomotors, with the pull cables in green. Every structural part is 3D printed and the contact pads are cast silicone.
    (b)~One finger: a single cable drives metacarpophalangeal (MCP), proximal interphalangeal (PIP) and distal interphalangeal (DIP) flexion, with PIP and DIP tied by a coupling cable.
    (c)~The thumb: carpometacarpal (CMC) abduction acts through a short linkage, while CMC flexion, MCP and interphalangeal (IP) motion are driven by two cables that share routing, so the three flexion channels are coupled.}
    \label{fig:hardware}
\end{widefigure}

\subsection{Design overview}
\revised{\HandName{} is a modular, human-scale, tendon-driven anthropomorphic hand. It is designed around a single premise, that a cable-driven hand can be made both learnable in simulation and repairable in the field.}
\revised{It has five fingers and sixteen revolute joints, measures $198 \times 95 \times 53.5$\,mm, and weighs $374$\,g.}
\rtwo{Seven servomotors drive them, fifteen of the joints through cables and the thumb abduction joint directly through a short linkage.}
\revised{The hand is deliberately underactuated and proprioceptively minimal}\rtwo{, and the only sensing is at the seven motor encoders.}
\reportonlytext{It mounts to a robot arm through a flat wrist-side surface with four M3 threaded inserts.}
Every design decision below serves one of three goals, namely anthropomorphic capability across the GRASP taxonomy, a transmission simple enough to reproduce faithfully in simulation, and a fully 3D-printed structure that a user can reprint and repair without a supplier (Fig.~\ref{fig:exploded}).

\subsection{\rtwo{Kinematics and transmission}}
\label{sec:transmission}

\rtwo{\emph{Fingers.}}
\revised{The four fingers are identical and underactuated. A single cable drives all three joints of a finger, so one motor governs three degrees of freedom, namely \rtwo{MCP, PIP and DIP flexion (Fig.~\ref{fig:finger})}.}
Which joint moves at any instant is set by the mechanics rather than commanded directly.
Extension is passive, provided by return springs, and the spring at the MCP is deliberately softer than those at the distal joints.
\revised{When the cable is drawn in,} the MCP therefore yields first and the finger begins to close at its base; only once the MCP reaches its $1.57$\,rad ($90^{\circ}$) limit do the stiffer distal springs give way and the coupled PIP and DIP joints curl.
\revised{This staged closure is what lets an underactuated finger conform to an object. The proximal joint seats the object against the palm before the distal joints close over it. A single cable therefore produces a secure, self-adapting grasp across a range of object shapes.}
\revised{The PIP and DIP joints are additionally tied by a coupling cable so that they flex together.}
Each fingertip and the intermediate phalanx carry cast silicone pads for compliant, high-friction contact (Fig.~\ref{fig:finger}).

\rtwo{\emph{Thumb.}}
\revised{The thumb carries four joints, namely CMC abduction, CMC flexion, MCP and IP.}
Two cables and the abduction joint together determine three flexion joints, so the thumb is where the transmission departs from the one-cable-per-finger pattern of the fingers.
The CMC abduction joint is driven directly through a short linkage rather than a long cable, and swings the thumb across the palm over a $0$--$100^{\circ}$ ($1.75$\,rad) range.
The CMC flexion cable then flexes the thumb by about $55^{\circ}$ across the palm, and the flexor cable, spanning MCP and IP, curls it by a further $90^{\circ}$ toward the palm to produce opposition and closure.
\revised{Because these three channels share routing, they are mutually coupled. Swinging the thumb out pays out cable on both flexion channels, and CMC flexion couples into them with opposite signs.}
\rtwo{This three-way coupling is what gives the thumb its dexterity.}
The MCP and IP joints of the thumb are additionally coupled to one another, as PIP and DIP are on the fingers.
Fig.~\ref{fig:thumb} labels the four thumb joints.

\rtwo{\emph{Cables and return springs.}}
\revised{Actuation is by cable and extension is passive throughout the hand.
The pull cables are braided Kevlar/Vectran, chosen for high tensile strength and low stretch, and each motor winds its cable on a $9.0$\,mm spool.}
The travel from the fully open hand to a closed fist is $44.45$\,mm of cable on the finger channels and $16.57$--$20.64$\,mm on the thumb channels, all consistent with that single spool radius.
\revised{The cables are routed over pulleys whose positions are taken from the CAD model, and the routed cables are visible in green in Fig.~\ref{fig:exploded}.}
\revised{Extension is provided throughout by music-wire return springs. Their stiffness is graded, softer at the MCP and stiffer at the distal joints, which sets the order in which the joints close.}
Because flexion is powered and extension is sprung, the resting posture of the hand is the fully open pose.

\subsection{\rtwo{Actuation, sensing and grasp capability}}
\label{sec:electronics}

\begin{reportonly}
\rtwo{\emph{Electronics and low-level control.}}
\rtwo{The six cables and the abduction linkage are driven by seven} Feetech HLS3606M serial-bus servomotors, each with an integrated magnetic encoder and about $0.59$\,N\,m of stall torque, packaged in a printed servo frame behind the palm.
An onboard ESP32-S3 microcontroller drives all seven servos directly over the serial bus; the board needs only two external connections, a $24$\,V DC power input and a single USB-C link to the host that carries both commands and feedback.
At the lowest level each servo runs its own position controller with a trapezoidal motion profile\rtwo{, and shows no measurable overshoot on any channel}.
\end{reportonly}

\begin{confonly}
\rtwo{\emph{Electronics and low-level control.}}
\rtwo{The six cables and the abduction linkage are driven by seven} Feetech HLS3606M serial-bus servomotors, each with an integrated magnetic encoder and about $0.59$\,N\,m of stall torque, and an onboard ESP32-S3 microcontroller drives all seven over a single serial bus.
Each servo runs its own position controller with a trapezoidal motion profile\rtwo{, and shows no measurable overshoot on any channel}.
\end{confonly}

\rtwo{\emph{Sensing and the control interface.}}
\label{sec:sensing}
The only proprioception on the hand is the seven motor encoders, which report $16$-bit positions over the serial bus.
We take this constraint as a design commitment: exposing exactly the quantities the hardware can measure defines an honest observation space for learning and removes the temptation to train on joint states the robot cannot provide at run time.

\rtwo{\emph{Grasp diversity and fingertip force.}}
\label{sec:graspdiversity}
\rtwo{Despite driving sixteen joints with only seven motors, the hand performs all $33$ grasp types of the GRASP taxonomy, with a single hardware configuration and without any change of hardware between grasps (Fig.~\ref{fig:taxonomy}).}
This breadth follows from the thumb's ability to oppose the four fingers over its $100^{\circ}$ abduction range, which supplies the stable thumb-against-fingers torque balance that most human tools assume.

\label{sec:force}
\rtwo{The hand is not merely posturally expressive but delivers usable grasping force.}
Measured at the tip, each finger and the thumb exert approximately $12$\,N, and the hand completes a full open-and-close cycle at about $1.2$\,Hz.

\begin{reportonly}
\subsection{\texorpdfstring{\rtwo{Repairability, durability and fabrication}}{Repairability, durability and fabrication}}
\label{sec:repair}

\rtwo{\emph{Repairability.}}
The entire structure of \handname{} is 3D printed, and the design files, the firmware and the control stack are released in full.
\rtwo{In general the hand is among the parts of a robot most exposed to damage.
A damaged part is reprinted from the released files rather than reordered.}
\revised{The bill of materials is dominated by the seven servomotors at \$208.81, roughly two-thirds of the total, with the ESP32-S3 board adding under \$10. The remainder goes to music-wire springs, Kevlar/Vectran cable, steel bearings and pins, fasteners and a few grams of filament.}

\rtwo{\emph{Durability.}}
\label{sec:robustness}
We characterise durability along three axes, following the protocol established for low-cost learning hands \cite{shaw2023leaphand}.

\emph{Endurance.}
Cycling the whole hand between the fully open and fully closed poses, the transmission survives more than $400{,}000$ full-actuation cycles without cable failure or loss of range, which at the nominal $1.2$\,Hz cycle rate corresponds to over $90$\,h of continuous articulation.
This exceeds the actuation a policy accumulates over a long training-and-deployment campaign, so cable wear is not a practical limit on the platform.

\emph{Cable strength.}
The actuation cable is the highest-stress element of the transmission, and it carries far more tension than the fingertip ever exerts.
\revised{The fingertip acts at a long moment arm about each joint while the cable pulls at the small $9.0$\,mm winding radius, so the transmission trades force for range.} The cable tension is larger than the fingertip force by the ratio of those moment arms, so \revised{an approximately} $12$\,N fingertip force corresponds to cable tensions of order $100$\,N.
The relevant bound is therefore the maximum tension the actuator can produce, which is set by the servo's $0.59$\,N\,m stall torque acting on the $9.0$\,mm spool, about $66$\,N.
The braided Kevlar/Vectran pull cable has a tensile strength of $69$\,kg, about $680$\,N, roughly a tenfold margin over that worst-case actuator tension.
\revised{Under overload the compliant silicone pads and the printed phalanges absorb the load and, if necessary, yield first, and a yielded phalanx is reprinted.}

\emph{Repeatability.}
Commanding a fixed target posture from the open pose across repeated closures, the hand returns to the same configuration to within $3$\,mm of cable excursion, comfortably inside the $44.45$\,mm finger stroke.
This repeatability is what allows a policy trained around a nominal grasp to rely on the hand reaching the same starting configuration at every reset.

\rtwo{\emph{Fabrication and assembly.}}
\label{sec:fabrication}
\revised{Every structural part is printed in \rtwo{polylactic acid}, chosen for stiffness and dimensional accuracy, on a Bambu Lab X1C with a $0.4$\,mm nozzle at a $0.2$\,mm layer height with tree supports. Any fused deposition modelling (FDM) printer with a bed of at least $200\times200$\,mm is sufficient.}
The compliant fingertip and palm pads are cast from soft silicone in printed two-part moulds that are released alongside the structural parts.
The non-printed parts are the seven servos, the Kevlar/Vectran pull cables, the music-wire return springs, and a small set of steel bearings, pins and self-tapping screws with brass heat-set inserts for the mounting face.
A complete set of structural parts prints in about $13$\,h on a single machine, and an individual finger in roughly $45$\,min, depending on layer height and print settings.
\end{reportonly}

\section{Tendon-Driven Simulation Model}
\label{sec:model}
The hand model is written for MuJoCo \cite{todorov2012mujoco} and released through MuJoCo Menagerie \cite{menagerie2022}.
It contains 16 hinge joints, 20 spatial tendons and 7 actuators (Fig.~\ref{fig:simmodel}).

\begin{widefigure}[!t]
    \centering
    \includegraphics[width=0.92\linewidth]{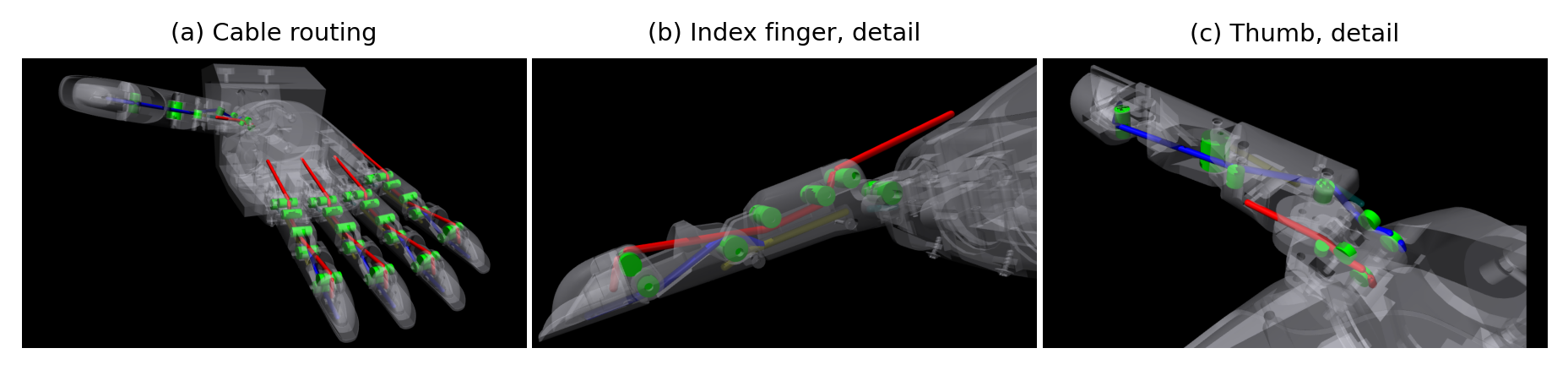}
    \caption{Tendon-driven hand model in MuJoCo. \rtwo{(a)~The whole hand,} in the extended pose with the link surfaces semi-transparent. Motor-driven actuation cables (red) run from palm anchor sites to the distal phalanges; passive coupling cables (blue) and extension springs (yellow, cyan) are spatial tendons as well; the green cylinders are the wrapping geometries, posed from the CAD model, that reproduce the physical pulley routing. (b)~One finger in isolation, showing the complete cable path from palm anchor to distal phalanx. \rtwo{(c)~The thumb in isolation: its two cables cross the abduction axis over the pulley cluster on the carpometacarpal base link, after which the CMC flexion cable acts on the CMC joint and the flexor cable continues to the MCP and IP joints.}}
    \label{fig:simmodel}
\end{widefigure}

\subsection{Cable transmission as spatial tendons}
Each finger carries four spatial tendons.
The first is an actuation cable that runs from an anchor site on the palm, over three wrapping cylinders on the proximal link, to the distal phalanx.
\revised{The second is a coupling cable that ties the PIP joint to the DIP joint.}
The remaining two are extension springs.
\revised{The thumb carries an actuation cable for CMC flexion, a flexor cable spanning its MCP and IP joints, and two springs.}
Springs are modelled as tendons with a rest length and a stiffness taken from the physical spring specifications.
The MCP springs use $352$\,N/m with a rest length of $11.376$\,mm, the distal springs $4000$\,N/m with $21.336$\,mm, and the thumb CMC spring $1897$\,N/m with $13$\,mm.

The routing geometry is what makes the model reproduce the physical transmission rather than merely approximate it.
A spatial tendon in MuJoCo wraps around a geometry only if the path is told which side to pass on.
\rtwo{We therefore place a cylinder at each physical pulley location, taken from the CAD model, with radii from $1.7$ to $4.5$\,mm and $2.5$\,mm on most of them.}
\rtwo{The wrap side is then specified explicitly on all twenty-four wraps of the four finger cables, and on two of the eight thumb wraps; the remaining thumb wraps and the return springs leave it to the solver.
Without it, the solver is free to route a cable on the wrong side of a pulley, which reverses the sign of its moment arm.
The moment arms of the finger transmission are therefore a consequence of the routing geometry rather than free parameters.}
The mechanical coupling between PIP and DIP is additionally enforced with joint equality constraints, as is the thumb MCP--IP coupling.

\subsection{Actuation interface}
Six position actuators regulate tendon lengths, with a position gain $k_p = 10^4$.
The thumb CMC abduction is driven by a position actuator acting directly on its joint, with $k_p = 1$.
This matches the physical routing, in which that channel drives the joint through a short linkage rather than a long cable.
The seven-dimensional control input therefore corresponds one-to-one to the seven motor spools, and six \texttt{tendonpos} sensors plus one \texttt{jointpos} sensor expose exactly the quantities the encoders provide.
The calibrated command intervals are $[0.0585, 0.1104]$\,m for the four finger cables, $[0.0262, 0.0384]$\,m and $[0.0816, 0.1121]$\,m for the two thumb cables, and $[-0.1, 1.75]$\,rad for the abduction joint.

\subsection{Numerical considerations}
Massively parallel training on a graphics processing unit imposes constraints beyond fidelity.
We integrate with semi-implicit Euler at a $10$\,ms timestep with implicit integration of joint damping disabled, replace all mesh collision geometries by manually placed primitives, and curate contact pairs explicitly: contacts within a finger and between the palm and the proximal links are excluded, while the fingertip-to-object and fingertip-to-palm contacts the task requires are preserved.
The reflected rotor inertia, $J_r n^2 = 1.56\times10^{-3}$\,kg\,m$^2$ with $J_r = 0.371\times10^{-7}$\,kg\,m$^2$ the rotor inertia of one motor and $n = 205$ its gear ratio, is assigned as joint armature.

\ifdefined\icraversion
\section{Actuation Mapping and System Identification}
\label{sec:sysid}

\subsection{Bidirectional actuation map}
\label{sec:notation}
Let $\mathbf{q}\in\mathbb{R}^{16}$ collect joint angles from the extended pose,
$\boldsymbol{\ell}\in\mathbb{R}^{7}$ the simulator-side tendon lengths (with
thumb abduction in radians), $\mathbf{d}$ the hardware cable travel in
millimetres, and $u_i\in\{0,\ldots,65535\}$ the motor commands and encoder
readings. The CAD-derived linear map $\mathbf{d}=\mathbf{Aq}$ uses effective
winding radii. For each finger and the two thumb cables it is
\begin{align}
 d_{\mathrm f} &= 12.49q_{\mathrm{mcp}}+7.32q_{\mathrm{pip}}+9.00q_{\mathrm{dip}},
 \label{eq:fingermap}\\
 d_{\mathrm{th,cmc}} &= 2.50q_{\mathrm{abd}}+12.49q_{\mathrm{cmc}},
 \label{eq:thumbcmc}\\
 d_{\mathrm{th,flexor}} &= 2.50q_{\mathrm{abd}}-2.50q_{\mathrm{cmc}}
 +9.44q_{\mathrm{mcp}}+12.50q_{\mathrm{ip}}.
 \label{eq:thumbflexor}
\end{align}
Coefficients are cable excursion divided by joint excursion between CAD poses;
the $2.50$\,mm/rad terms are physical pulley radii. The negative term records
cable paid out as CMC flexion winds the distal cable onto its pulley. Six cable
channels use the $9.0$\,mm spool radius $r$, while thumb abduction is driven
directly:
\begin{align}
 \theta_i &= d_i/r, \label{eq:spool}\\
 u_i &= \left\lfloor
 \frac{\theta_i-\theta_i^{\min}}{\theta_i^{\max}-\theta_i^{\min}}65535
 \right\rceil . \label{eq:motormap}
\end{align}
The limits are the extrema of $\mathbf{Aq}$ over the joint-limit box. They give
$288.16^{\circ}$ of shaft travel on each finger, $100^{\circ}$ on abduction,
$104.13^{\circ}$ on thumb CMC flexion, and $[-15.28,247.15]^{\circ}$ on the
thumb flexor. The negative endpoint is a consequence of the thumb cross term,
not a calibration offset.

For a finger, the simulator and motor intervals give the affine bidirectional
map
\begin{equation}
 \ell_i(u_i)=\ell_i^{\max}-\frac{u_i}{65535}
 (\ell_i^{\max}-\ell_i^{\min}).
 \label{eq:servomap}
\end{equation}
The thumb instead requires the coupled joint configuration. We sweep its two
tendons in simulation and regress the three flexion angles on tendon lengths and
abduction,
\begin{equation}
 \hat{\mathbf q}_{\mathrm{th}}=\mathbf C
 [q_{\mathrm{abd}},\ell_1,\ell_2,1]^{\top},
 \label{eq:thumbfit}
\end{equation}
where
\begin{equation}
 \mathbf C=\begin{bmatrix}
 3.4\!\times\!10^{-4}&-78.09&0.19&2.977\\
 4.2\!\times\!10^{-3}&-11.37&-56.72&6.666\\
 4.5\!\times\!10^{-3}&-11.42&-56.89&6.687
 \end{bmatrix}.
 \label{eq:thumbcoef}
\end{equation}
The first-order fit reaches $R^2\geq0.9995$; quadratic terms do not improve it.

\begin{widefigure}[!t]
 \centering
 \includegraphics[width=\linewidth]{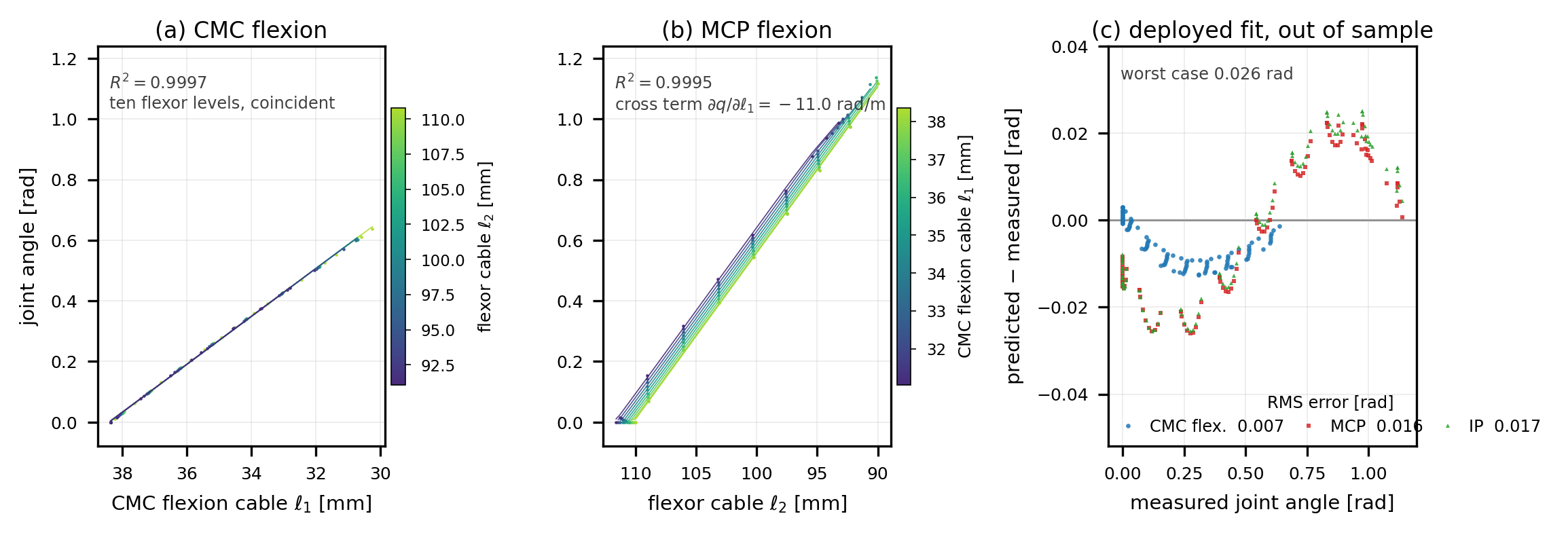}
 \caption{Thumb-map identification on a $10\times10$ tendon sweep. (a) CMC
 flexion depends primarily on its own cable. (b) MCP motion exposes the
 cross-coupling from the CMC cable; IP coincides through the joint constraint.
 (c) Out-of-sample residuals of \eqref{eq:thumbfit}. Solid lines are the linear
 fit.}
 \label{fig:thumbmap}
\end{widefigure}

Because \eqref{eq:thumbfit} is not invertible, encoder readings use two
first-order cable fits,
\begin{align}
 d_{\mathrm{th,cmc}} &= \alpha_1\ell_1+\beta_1+2.50q_{\mathrm{abd}},\nonumber\\
 d_{\mathrm{th,flexor}} &= \alpha_2\ell_2+\beta_2+2.50q_{\mathrm{abd}},
 \label{eq:thumbinv}
\end{align}
with $(\alpha_1,\beta_1)=(-977.2,37.52)$ and
$(\alpha_2,\beta_2)=(-1241.6,136.59)$. The measured hand has
abduction-to-flexion coupling that is nearly absent from the simulated sweep, so
the analytical $2.50$\,mm/rad contribution is removed for fitting and restored
at run time. The controller therefore evaluates only affine maps in both
directions; transmission nonlinearity remains in the routed tendon geometry.

\subsection{Validation}
Cable excursion from an extended hand to a fist is measured independently from
hardware encoder end stops and simulator tendon sensors. The four finger errors
are $0.04$, $0.32$, $1.40$, and $0.41$\,mm (at most $3.1\%$ of stroke).
Abduction is $6.7\%$ short of the servo excursion; the two thumb cable
excursions differ by $32.7\%$ and $31.2\%$. The CMC effective winding radius
agrees with CAD to $1.1\%$, whereas the flexor radius is $16.8\%$ low; the
runtime coupling term compensates the remaining abduction error.

A shared $134$\,s staircase command then exercises every channel over its full
range. On settled samples, the finger cables track within $0.29$--$0.45$\,mm
RMS ($0.6$--$1.0\%$ of full scale), and the worst thumb channel within
$0.63$\,mm RMS. Simulated rise times lead the hardware by $10$--$40$\,ms. The
largest structured residual, $7.1\%$ of full scale, is the thumb cross-coupling
represented only to first order.

\begin{widetable}[!t]
 \centering
 \footnotesize
 \begin{tabular}{lrrrr}
 \toprule
 & \multicolumn{2}{c}{Amplitude} & \multicolumn{2}{c}{$10$--$90\%$ rise time}\\
 \cmidrule(lr){2-3}\cmidrule(lr){4-5}
 Channel & Excursion $|\Delta|$ & RMS & Simulation & Hardware\\
 \midrule
 Thumb abduction [mrad] & $124.94$ & $26.79$ & $70$\,ms & $80$\,ms\\
 Thumb CMC cable [mm]   & $5.42$   & $0.11$  & $40$\,ms & $80$\,ms\\
 Thumb flexor [mm]      & $6.44$   & $0.63$  & $60$\,ms & $100$\,ms\\
 Index cable [mm]       & $0.04$   & $0.45$  & $90$\,ms & $120$\,ms\\
 Middle cable [mm]      & $0.32$   & $0.39$  & $90$\,ms & $120$\,ms\\
 Ring cable [mm]        & $1.40$   & $0.39$  & $90$\,ms & $120$\,ms\\
 Pinky cable [mm]       & $0.41$   & $0.29$  & $90$\,ms & $120$\,ms\\
 \bottomrule
 \end{tabular}
 \caption{Independent excursion error, settled tracking RMS, and median rise
 time under the shared staircase command.}
 \label{tbl:validation}
\end{widetable}

\else
\section{Actuation Mapping and System Identification}
\label{sec:sysid}

\subsection{Notation}
\label{sec:notation}
Channels are indexed by $i \in \mathcal{I} = \{1,\dots,7\}$, ordered as index, middle, ring and pinky cable, thumb abduction, \revised{thumb CMC flexion cable}, and thumb flexor cable.
Joints are indexed by $j \in \mathcal{J} = \{1,\dots,16\}$.
Four quantities appear throughout, and the actuation map of this section relates them.
\rtwo{The joint angle $q_j \in \mathbb{R}$ is taken as zero in the fully extended pose and increases with flexion, and we write $\mathbf{q} = (q_1,\dots,q_{16})$.}
Individual joints are named rather than numbered where that is clearer, so $q_{\mathrm{mcp}}$, $q_{\mathrm{pip}}$ and $q_{\mathrm{dip}}$ are the \rtwo{MCP, PIP and DIP} angles of one finger, while $q_{\mathrm{abd}}$, $q_{\mathrm{cmc}}$ and $q_{\mathrm{ip}}$ are the thumb abduction, CMC flexion and \rtwo{IP} angles.
The simulation-side command of channel $i$ is $\ell_i \in \mathbb{R}$, collected into $\boldsymbol{\ell} = (\ell_1,\dots,\ell_7)$.
For the six cable channels it is the length of the corresponding spatial tendon in metres, as reported by the tendon-length sensor of the model, and for the abduction channel, $i = 5$, it is the joint angle $q_{\mathrm{abd}}$ in radians, because that channel drives its joint directly.
The hardware-side actuation of channel $i$ is $d_i \in \mathbb{R}$, defined as the cable travel spooled in by its motor relative to the fully extended pose, in millimetres; it differs from $\ell_i$ by an offset and a sign, since a longer tendon in the model corresponds to less cable spooled in on the hand.
\mohit{The motor-side actuation of channel $i$ is $\theta_i \in \mathbb{R}$, the rotation of the output shaft of its motor away from the fully extended pose, and $\boldsymbol{\theta} = (\theta_1,\dots,\theta_7)$; for the six cable channels it is $d_i$ divided by the spool radius, and for the abduction channel it is the joint angle itself.}
Finally, $u_i \in \{0,1,\dots,65535\}$ is the 16-bit position command sent to motor $i$, and also the value its encoder returns.

\subsection{\rtwo{The joint-to-actuation model}}
\label{sec:jta}
\rtwo{The hand carries its own model of how joint angles produce cable travel, motor rotation and finally a motor command.
It is a property of the mechanism, taken from the CAD model.}

\rtwo{The model is linear.}
\revised{Its coefficients are effective winding radii, in millimetres of cable travel per radian of joint rotation.}
\begin{mohitblock}
A winding radius is a length, so a magnitude $|c_{\bullet}|$ is positive and the sign carries the direction of the routing: a positive entry means the joint takes cable up on that channel as it flexes, and a negative one means it pays cable out.
Every finger entry is positive, because a finger cable can only take up as the finger closes.
The map has exactly one negative entry, on the thumb, and it is explained where it arises.

The magnitudes are taken from the CAD model rather than fitted to measurements.
One joint is swept at a time with the others held extended, the routed cable path is evaluated at the two ends of that joint's travel, and the coefficient is the cable travel $\Delta d_j$ taken up over the angular range,
\begin{equation}
  c_j = \frac{\Delta d_j}{\Delta q_j}.
  \label{eq:windingcoeff}
\end{equation}
Linearity is an assumption of the model and not a result of this construction: two poses fix a slope, and the path length is taken to vary linearly between them.
\end{mohitblock}

\mohit{\emph{Fingers.} Each finger carries three joints, MCP, PIP and DIP, on a single cable driven by a single motor, so one actuation variable has to account for all three rotations at once:}
\begin{equation}
  d_{\mathrm{f}} = c_{\mathrm{mcp}} q_{\mathrm{mcp}} + c_{\mathrm{pip}} q_{\mathrm{pip}} + c_{\mathrm{dip}} q_{\mathrm{dip}},
  \label{eq:fingermap}
\end{equation}
with $(c_{\mathrm{mcp}}, c_{\mathrm{pip}}, c_{\mathrm{dip}}) = (12.49, 7.32, 9.00)$\,mm/rad, where the subscript $\mathrm{f}$ stands for any one of the four fingers.
\begin{reportonly}
\begin{mohitblock}
Fig.~\ref{fig:windingcoeff} shows the construction for the MCP joint, whose range is $[0, 90]^{\circ}$.
The red portion of the tendon is the stretch whose length does not change as the flex joint moves, so it cancels in the difference and only the remaining stretch matters.
That stretch is drawn in blue in the extended pose and in cyan in the fully flexed pose, and the difference between the two lengths, taken from CAD, is the cable take-up $\Delta d_{\mathrm{mcp}} = 19.62$\,mm over the $90^{\circ}$ of travel, so \eqref{eq:windingcoeff} gives $c_{\mathrm{mcp}} = 19.62 / (\pi/2) = 12.49$\,mm/rad.
The PIP and DIP coefficients follow from the same two-pose evaluation on their own joints.
\end{mohitblock}
\end{reportonly}

\begin{reportonly}
\begin{figure}[!htbp]
    \centering
    \includegraphics[width=0.62\textwidth]{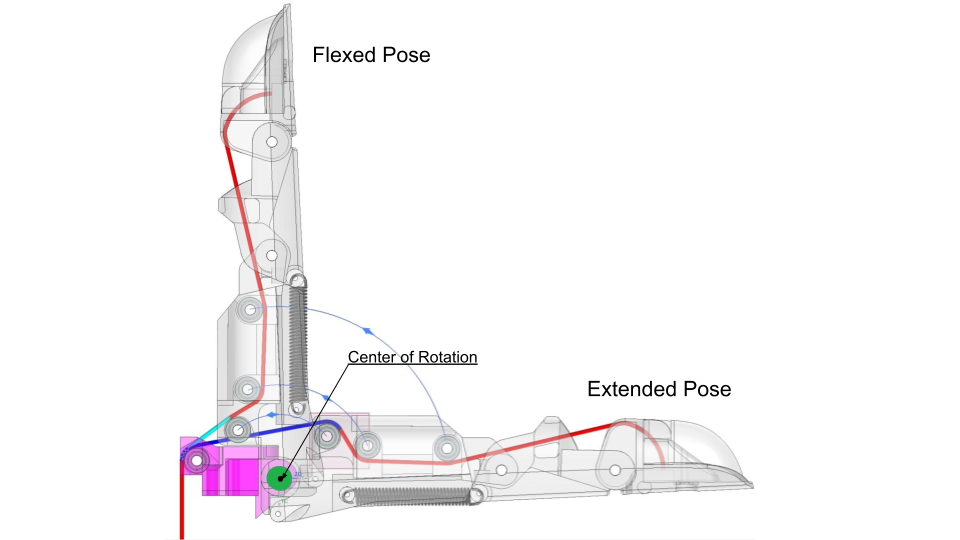}
    \caption{\mohit{Construction of a winding coefficient, shown for the MCP joint of one finger. The routed cable path is evaluated in the extended pose and in the fully flexed pose, and the coefficient is the difference in path length over the $90^{\circ}$ of joint travel. Red marks the stretch of tendon that is unaffected by the joint, blue the remaining stretch in the extended pose and cyan the same stretch in the fully flexed pose.}}
    \label{fig:windingcoeff}
\end{figure}
\end{reportonly}

\mohit{\emph{Thumb.} The thumb carries four joints, abduction, CMC flexion, MCP and IP, on three channels, so it cannot be written as one independent row per channel in the way a finger can.
Abduction drives its own joint directly through a linkage.
The other two channels are cables. Both cross the abduction axis, and the flexor cable crosses the CMC axis as well, which is where the cross terms come from:}
\begin{align}
  d_{\mathrm{th,cmc}} &= 2.50\, q_{\mathrm{abd}} + 12.49\, q_{\mathrm{cmc}}, \label{eq:thumbcmc}\\
  d_{\mathrm{th,flexor}} &= 2.50\, q_{\mathrm{abd}} - 2.50\, q_{\mathrm{cmc}} + 9.44\, q_{\mathrm{mcp}} + 12.50\, q_{\mathrm{ip}}. \label{eq:thumbflexor}
\end{align}
The abduction term enters both thumb channels with the same sign and magnitude, expressing that swinging the thumb out takes up cable on both; the CMC flexion term enters the two channels with opposite signs.
\begin{reportonly}
\begin{mohitblock}
Three of these entries do not need \eqref{eq:windingcoeff} at all, because the cable wraps a pulley of known radius.
A cable wrapping a pulley of radius $r$ through an angle $\theta$ takes up an arc $r\theta$, so its winding radius is the pulley radius itself.
Both thumb cables pass over the abduction pulley, of radius $2.50$\,mm, which fixes the abduction coefficient they share, and the flexor cable crosses the CMC pulley of the same radius, which fixes the magnitude of the $-2.50$\,mm/rad entry of \eqref{eq:thumbflexor}.
The three remaining coefficients, $12.49$\,mm/rad on CMC flexion and $9.44$ and $12.50$\,mm/rad on MCP and IP, come from the two-pose evaluation as for the fingers.

The sign of that entry is the one place in the map where a joint pays cable out rather than taking it up.
The flexor cable passes over the CMC pulley on its way to the distal joints, so flexing that joint winds the cable onto the pulley and feeds it forward.
Cable arriving at the distal joints this way has to be matched by cable released at the motor, or the MCP and IP angles would change as the CMC joint closes.
Holding the distal thumb posture through a CMC closure therefore costs $2.50$\,mm of released cable per radian, and that is what the negative entry records.
\end{mohitblock}
\end{reportonly}
\begin{confonly}
The $2.50$\,mm/rad entries are the corresponding pulley radii; the remaining coefficients follow from \eqref{eq:windingcoeff}. The negative CMC term records cable paid out to hold the distal thumb posture as CMC flexion winds cable onto its pulley.
\end{confonly}
\mohit{Collecting the seven rows gives $\mathbf{d} = \mathbf{A}\mathbf{q}$ with $\mathbf{A} \in \mathbb{R}^{7\times16}$ of rank seven, block diagonal apart from the thumb, whose three rows share the abduction and CMC flexion columns.}
\mohit{The abduction row is the identity on $q_{\mathrm{abd}}$ and is dimensionless, because that channel drives its joint through a linkage rather than a spool; the six cable rows carry millimetres per radian.}
\mohit{The map has no constant term, so the fully extended pose is the origin of both spaces by construction, and a change of angular unit passes through it unchanged.}

\begin{mohitblock}
Each of the six cable channels winds on a spool of radius $r = 9.0$\,mm, so the shaft rotation is
\begin{equation}
  \theta_i = d_i / r,
  \label{eq:spool}
\end{equation}
while the abduction channel takes $\theta_5 = q_{\mathrm{abd}}$ directly.
The firmware accepts a 16-bit position per channel, so each is normalised affinely over its own excursion,
\begin{equation}
  u_i = \left\lfloor \frac{\theta_i - \theta_i^{\min}}{\theta_i^{\max} - \theta_i^{\min}}\; 65535 \right\rceil .
  \label{eq:motormap}
\end{equation}
The endpoints here are not calibration constants.
They are the extrema of the same linear map over the box of admissible joint angles $\mathcal{Q} = \prod_j [\,q_j^{\min}, q_j^{\max}\,]$,
\begin{equation}
  \theta_i^{\min} = \frac{1}{r}\min_{\mathbf{q}\in\mathcal{Q}} (\mathbf{A}\mathbf{q})_i,
  \qquad
  \theta_i^{\max} = \frac{1}{r}\max_{\mathbf{q}\in\mathcal{Q}} (\mathbf{A}\mathbf{q})_i,
  \label{eq:actrange}
\end{equation}
and because $\mathbf{A}$ is linear each extremum is attained at a vertex of $\mathcal{Q}$, in a posture where every joint sits on one of its own limits.
With the joint box of this hand, $q_{\mathrm{abd}} \in [0, 100]^{\circ}$, $q_{\mathrm{cmc}} \in [0, 55]^{\circ}$ and every remaining flexion joint in $[0, 90]^{\circ}$, this gives an excursion of $[0, 288.16]^{\circ}$ of shaft rotation for each of the four finger cables, $[0, 100]^{\circ}$ for abduction, $[0, 104.13]^{\circ}$ for the thumb CMC flexion cable and $[-15.28, 247.15]^{\circ}$ for the thumb flexor cable.

The lower limit of the flexor channel is the one entry of that list which is not zero, and it follows from the cross terms rather than from a convention.
The extended pose is not the extreme of that channel: by \eqref{eq:thumbflexor} the flexor pays out further as the CMC joint flexes, so the minimum is attained at the vertex $q_{\mathrm{cmc}} = 55^{\circ}$ with the remaining thumb joints at zero, where the model gives $-2.50 \times 0.960 = -2.40$\,mm of cable and hence $-15.28^{\circ}$ of shaft rotation.
The hand can therefore reach a posture in which the flexor spool has unwound past its extended-pose reading, and a controller that clamped that channel below at zero would block full CMC flexion.
This is the one place where the thumb coupling is visible in a quantity that can be checked without instrumenting the hand, \rtwo{and the value the hand is commanded against agrees with \eqref{eq:actrange} to $1.1\times10^{-3}$ degrees.
Enumerating the $2^{16}$ vertices of $\mathcal{Q}$ reproduces the three thumb limits to better than $10^{-3}$ degrees and the four finger limits to $3.7\times10^{-2}$ degrees, the latter gap being the rounding of the winding coefficients to the four decimal places quoted in \eqref{eq:fingermap}.}
A second consequence of \eqref{eq:motormap} is that the command quantisation can be ignored: the finger excursion of $288.16^{\circ}$ over $65535$ counts is $0.69$\,$\mu$m of cable per count, \rtwo{between two and three orders of magnitude below the tracking residuals of the transmission.}
\end{mohitblock}

\subsection{The actuation map}
\label{sec:actmap}
\rtwo{The simulator is commanded in $\boldsymbol{\ell}$ and the hardware in $\mathbf{u}$, and the controller crosses between them twice per step, forwards to turn a policy output into a motor command and backwards to turn an encoder reading into an observation.
On the four finger channels the crossing is a single affine map.
On the three thumb channels it is not, and the missing link into \eqref{eq:thumbcmc} has to be identified.
The two families are therefore treated separately.}

\rtwo{\emph{Finger channels.}}

\rtwo{A finger is driven by a single cable, so $\ell_i$ and $d_i$ are two readings of the same cable, one taken in the model and one taken at the spool.
Their difference is an offset and a sign fixed by the routing, independent of posture, so the relation between the simulation command and the motor command is affine and both of its endpoints are already known.
The command endpoints are the excursion limits of \eqref{eq:actrange}.
The length endpoints $\ell_i^{\min}$ and $\ell_i^{\max}$ are obtained by sweeping actuator $i$ across its \texttt{ctrlrange} in simulation and recording the extreme values of its tendon-length sensor.
Matching one interval to the other gives the whole map,}
\begin{equation}
  \ell_i(u_i) = \ell_i^{\max} - \frac{u_i}{65535}\left(\ell_i^{\max} - \ell_i^{\min}\right).
  \label{eq:servomap}
\end{equation}
The map is decreasing because spooling cable in shortens a flexion tendon, so $u_i = 0$ gives $\ell_i = \ell_i^{\max}$ and $u_i = 65535$ gives $\ell_i = \ell_i^{\min}$.
The abduction channel uses the increasing convention instead.
\rtwo{Both directions are \eqref{eq:servomap}, read one way or solved the other.}

\rtwo{Anchoring the two intervals at their endpoints makes the map exact at the extremes but leaves its slope to be inherited rather than measured.
The three available estimates of the finger excursion do not agree.
The hand travels $44.45$\,mm of cable from open to fist (Sec.~\ref{sec:transmission}), \eqref{eq:actrange} gives $45.26$\,mm, and the simulated sweeps give $45.72$--$46.21$\,mm across the four fingers.
Equation \eqref{eq:servomap} absorbs the spread as a gain of $1.0$--$2.1\%$ on the simulation side.}

\rtwo{\emph{Thumb channels.}}

\rtwo{The thumb admits no such shortcut.
Two cables and the abduction joint together determine three flexion joints, and the abduction linkage pays out cable on both flexion channels, so $d$ depends on $q_{\mathrm{abd}}$ as well as on the two cable lengths.
The joint angles that \eqref{eq:thumbcmc} and \eqref{eq:thumbflexor} consume therefore have to be estimated from the simulation command.}
We drive the simulated hand over its thumb workspace, log the two cable lengths, the abduction angle and the resulting joint angles, and regress the joints on the inputs,
\begin{equation}
  \hat{\mathbf{q}}_{\mathrm{th}}
  = \mathbf{C}\,\bigl[\,q_{\mathrm{abd}},\; \ell_{1},\; \ell_{2},\; 1\,\bigr]^{\!\top},
  \label{eq:thumbfit}
\end{equation}
where $\hat{\mathbf{q}}_{\mathrm{th}} = (\hat{q}_{\mathrm{cmc}}, \hat{q}_{\mathrm{mcp}}, \hat{q}_{\mathrm{ip}}) \in \mathbb{R}^{3}$ are the estimated thumb flexion angles in radians, $\ell_1 = \ell_6$ and $\ell_2 = \ell_7$ are the CMC flexion and flexor cable lengths in metres, $q_{\mathrm{abd}}$ is in radians, and $\mathbf{C} \in \mathbb{R}^{3 \times 4}$ is the ordinary least squares solution over the logged sweep,
\begin{equation}
  \mathbf{C} =
  \begin{bmatrix}
     3.4\!\times\!10^{-4} & -78.09 &   0.19 & 2.977 \\
     4.2\!\times\!10^{-3} & -11.37 & -56.72 & 6.666 \\
     4.5\!\times\!10^{-3} & -11.42 & -56.89 & 6.687
  \end{bmatrix},
  \label{eq:thumbcoef}
\end{equation}
The four columns of $\mathbf{C}$ carry units of rad/rad, rad/m, rad/m and rad respectively.
\rtwo{The sweep and the fit are shown in Fig.~\ref{fig:thumbmap}.}
\rtwo{That a first-order model suffices is a measured property of the transmission rather than a modelling convenience.
A plain bilinear map in the two cable lengths already attains $R^2 \ge 0.9995$ on the sweeps, and adding quadratic terms does not improve it.
The residual minimised in \eqref{eq:thumbfit} is taken against the joint angles the simulator itself reports, so the cable travel that \eqref{eq:thumbcmc} and \eqref{eq:thumbflexor} return from those angles is the reference throughout.}

\begin{widefigure}[!t]
    \centering
    \includegraphics[width=\linewidth]{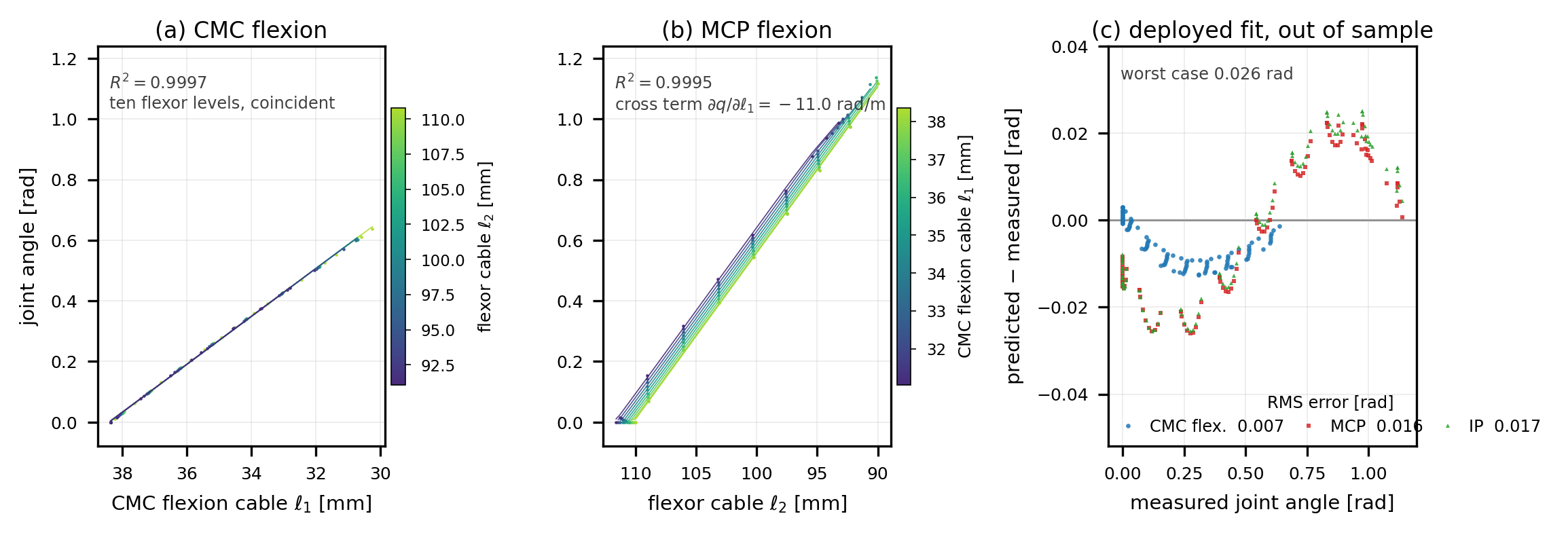}
    \caption{\rtwo{The first stage of the thumb map, measured on the released $10\times10$ sweep of the two thumb actuators with abduction held fixed. Both cable axes are reversed, so the cable shortens and the thumb closes from left to right, and the colour bar beside each panel gives the length of the other cable, one shade per level of the sweep. (a)~The CMC flexion joint against its own cable. The ten flexor levels coincide, so this joint is set by one cable alone. (b)~The MCP joint against the flexor cable. The ten levels of the CMC flexion cable are displaced from one another, and that displacement is the cross term the fit has to carry. The ten curves converge at the extended end, where the joint rests at its limit and the CMC flexion cable no longer moves it; the spread across the ten levels is $2\times10^{-4}$\,rad there against about $8\times10^{-2}$\,rad over the rest of the sweep, and that saturation is one thing a first-order model cannot represent. The IP joint is tied to MCP by an equality constraint and its curve coincides, so it is not drawn twice. Solid lines in both panels are the bilinear fit. (c)~Residuals of the coefficients \eqref{eq:thumbcoef} that the deployed controller runs, scored on this sweep. They were identified on a different capture in which the abduction joint was moving, so this is an out-of-sample check.}}
    \label{fig:thumbmap}
\end{widefigure}

\rtwo{The forward direction is now closed.
Given a commanded $\boldsymbol{\ell}$, \eqref{eq:thumbfit} returns $\hat{\mathbf{q}}_{\mathrm{th}}$, \eqref{eq:thumbcmc} and \eqref{eq:thumbflexor} return the two cable travels and \eqref{eq:spool} and \eqref{eq:motormap} return the commands.
This composition is what the controller evaluates at every control step.}

\rtwo{The inverse direction cannot reuse it, because $\mathbf{C}$ is $3 \times 4$ and \eqref{eq:thumbfit} is not invertible.
A separate pair of first-order fits of the same sweeps is used instead, one per cable,}
\begin{align}
  d_{\mathrm{th,cmc}} &= \alpha_1 \ell_1 + \beta_1 + 2.50\, q_{\mathrm{abd}}, \nonumber\\
  d_{\mathrm{th,flexor}} &= \alpha_2 \ell_2 + \beta_2 + 2.50\, q_{\mathrm{abd}},
  \label{eq:thumbinv}
\end{align}
where $\alpha_1, \alpha_2$ are gains in millimetres of cable travel per metre of tendon length and $\beta_1, \beta_2$ are offsets in millimetres, with $(\alpha_1,\beta_1) = (-977.2,\, 37.52)$ and $(\alpha_2,\beta_2) = (-1241.6,\, 136.59)$.
Given the measured $d_{\mathrm{th,cmc}}$, $d_{\mathrm{th,flexor}}$ and $q_{\mathrm{abd}}$, \eqref{eq:thumbinv} is solved for $\ell_1$ and $\ell_2$ after subtracting the abduction term.

\rtwo{The abduction term in \eqref{eq:thumbinv} is written analytically rather than fitted, and the reason is a discrepancy between the simulator and the mechanism.}
The first column, however, is numerically negligible.
It stays below $5\times10^{-3}$\,rad/rad, so the entire $1.75$\,rad abduction range moves the flexion joints by less than $0.01$\,rad.
The simulated thumb has almost no abduction-to-flexion coupling, whereas \eqref{eq:thumbcmc} and \eqref{eq:thumbflexor} assert $2.50$\,mm of cable travel per radian of abduction.
\rtwo{The two fits of \eqref{eq:thumbinv} are therefore taken after that contribution has been removed from the reference, and it is added back at run time.
The mapping compensates in software for a coupling that the simulator does not reproduce.}

\rtwo{\emph{Both directions.}}

\rtwo{Stages \eqref{eq:spool} and \eqref{eq:motormap} invert in closed form for every channel,
\begin{equation}
  \theta_i = \theta_i^{\min} + \frac{u_i}{65535}\left(\theta_i^{\max} - \theta_i^{\min}\right),
  \qquad d_i = r\,\theta_i,
  \label{eq:inversechain}
\end{equation}
after which \eqref{eq:thumbinv} returns the two thumb cable lengths and \eqref{eq:servomap} the four finger ones.}
\rtwo{A compact seven-value joint representation is also available, one value per finger and three for the thumb, in which the joints driven by a common cable are assigned a common angle.}
\mohit{That representation is a section of the joint space on which $\mathbf{A}$ is one-to-one, and it is the form in which hand postures are specified in practice.}

We emphasise that every stage of this chain is linear or affine.
The nonlinearity of the physical transmission is carried by the routing geometry of the model itself, not by the interface.

\subsection{Kinematic validation}
\label{sec:kinematic}
The kinematic test compares two independent estimates of one quantity, the cable excursion between the fully extended hand and a fist.
On the hardware this follows from the measured end-stop encoder excursions and the $9.0$\,mm spool radius; in the model it follows from the CAD-derived routing geometry, and is measured from the tendon-length sensors at the two end stops.
Neither number is fitted to the other.

The four finger cables agree to $0.04$, $0.32$, $1.40$ and $0.41$\,mm (Table~\ref{tbl:validation}).
The largest of these is $3.1\%$ of a $44.45$\,mm excursion, and the others are below $1\%$.
The abduction joint reaches $1.746$\,rad against a servo travel of $1.871$\,rad, a $6.7\%$ shortfall that is explained by the joint limit of the model being slightly tighter than the mechanical travel.
The two thumb cable channels are the outliers, at $32.7\%$ and $31.2\%$.
The model reproduces neither the full CMC flexion excursion nor the flexor excursion.
The fit of \eqref{eq:thumbfit} localises the cause.
For the CMC flexion cable the model realises an effective winding radius of $12.64$\,mm/rad, which agrees with the CAD value in \eqref{eq:thumbcmc} to $1.1\%$.
The flexor cable, however, realises $18.25$\,mm/rad against the $21.94$\,mm/rad of \eqref{eq:thumbflexor}, a shortfall of $16.8\%$.
The missing abduction coupling accounts for the rest.
Since the mapping restores the abduction term in software, what survives to the hardware is the flexor-radius error.

\subsection{Dynamic validation}
\label{sec:dynamic}
One staircase command stream drives both systems for $134$\,s, exercising all seven channels over their whole range, one channel at a time\reportonlytext{ (Fig.~\ref{fig:validation})}.
\rtwo{Two properties are asked of the model.
Each channel should settle to the same value as the hardware once the mapping is applied, and it should take a comparable time to get there.
Both hold.
Residuals are evaluated on settled samples only, meaning that no command has changed for $0.4$\,s and neither trace has moved by more than $0.5\%$ of its range over the preceding $100$\,ms, which retains $60\%$ of the $6643$ samples of the run.}
The four finger cables track to $0.29$--$0.45$\,mm RMS, \rtwo{that is} $0.6$--$1.0\%$ of full scale, \rtwo{and the thumb flexor cable, the worst channel, to $0.63$\,mm RMS.}
\rtwo{The $10$--$90\%$ rise time is $90$\,ms in simulation against $120$\,ms on the hardware for every finger, and $40$--$70$\,ms against $80$--$100$\,ms on the thumb channels, so the simulated actuator is consistently $10$--$40$\,ms faster than the physical servo.}
\rtwo{The one channel that does not agree in amplitude is the thumb flexor while the CMC flexion cable is swept, where the hardware moves and the model barely does, and the flexor residual reaches $7.1\%$ of full scale.
That is the coupling which \eqref{eq:thumbfit} carries only to first order.}

\begin{widetable}[!t]
    \centering
    \begin{tabular}{lrrrr}
    \toprule
    & \multicolumn{2}{c}{Amplitude} & \multicolumn{2}{c}{$10$--$90\%$ rise time} \\
    \cmidrule(lr){2-3}\cmidrule(lr){4-5}
    Channel & Excursion $|\Delta|$ & Tracking RMS & Simulation & Hardware \\
    \midrule
    Thumb CMC abduction (joint) [mrad] & $124.94$ & $26.79$ & $70$\,ms & $80$\,ms \\
    Thumb CMC flexion cable [mm]       & $5.42$   & $0.11$  & $40$\,ms & $80$\,ms \\
    Thumb flexor cable [mm]            & $6.44$   & $0.63$  & $60$\,ms & $100$\,ms \\
    Index cable [mm]                   & $0.04$   & $0.45$  & $90$\,ms & $120$\,ms \\
    Middle cable [mm]                  & $0.32$   & $0.39$  & $90$\,ms & $120$\,ms \\
    Ring cable [mm]                    & $1.40$   & $0.39$  & $90$\,ms & $120$\,ms \\
    Pinky cable [mm]                   & $0.41$   & $0.29$  & $90$\,ms & $120$\,ms \\
    \bottomrule
    \end{tabular}
    \caption{\rtwo{Sim-to-real validation of the transmission model. \emph{Excursion} is the difference between the model and the hardware in the cable travel from the extended hand to a fist, measured independently on each system. \emph{Tracking RMS} is the residual between the simulated sensor and the encoder-derived actuation value over a $134$\,s run in which one command stream drives both systems, evaluated on settled samples only. The rise times are medians over $72$--$177$ matched steps per channel. Cable channels are reported in millimetres and the abduction joint in milliradians. Simulated rise times are reconstructed from the integrator step count, since the acquisition loop advances the simulator at about half real time; the $10$\,ms simulation timestep and the $20$\,ms hardware sampling period bound their resolution.}}
    \label{tbl:validation}
\end{widetable}

\begin{reportonly}
\begin{widefigure}[!t]
    \centering
    \includegraphics[width=0.80\linewidth]{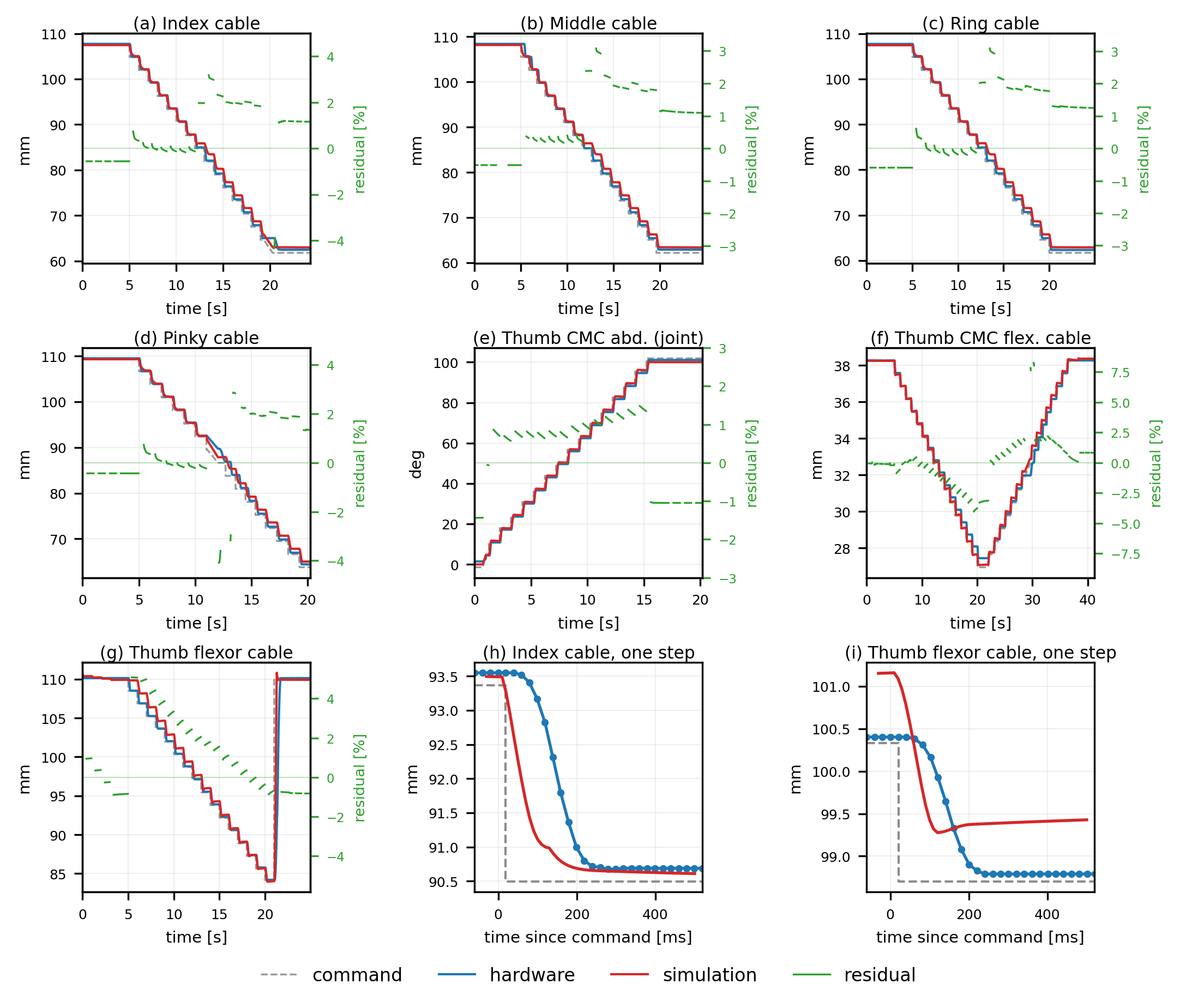}
    \caption{Sim-to-real validation of the transmission model. One staircase command stream (grey, dashed) drives the simulator (red) and the physical hand (blue). (a)--(g)~The seven channels. The sweep commands one channel at a time, so each panel is cropped to the window in which its own channel is stepped, with five seconds either side that show the converged state before and after. Each panel also carries, on the right-hand axis, the residual of that same channel over the same window, normalised by its calibrated range. Only settled samples of the residual are drawn, so its gaps are the discarded transients. The finger channels stay within $\pm 2\%$. (h),~(i)~One resolved step of a finger channel and of a thumb channel. Each system is plotted against its own time base, because the acquisition loop advances the integrator by $10$\,ms per logged row while the wall clock advances $20$\,ms. The simulated actuator leads the servo by $10$--$40$\,ms, and the thumb channel undershoots before creeping back, whereas the hardware runs a trapezoidal motion profile and does not overshoot.}
    \label{fig:validation}
\end{widefigure}
\end{reportonly}

\fi
\section{Learning In-Hand Cube Rotation}
\label{sec:rotatez}
On the identified model we train a policy that continuously rotates a cube about the vertical axis while maintaining a stable grasp, a task widely used to evaluate sustained rotational manipulation \cite{andrychowicz2020learning, openai2019rubik, qihaozhi-arllegro}.
We formulate the task in MuJoCo Playground \cite{zakka2025mujoco} and train with Proximal Policy Optimization (PPO) \cite{schulman2017proximal} as implemented in Brax \cite{freeman2021brax}.
The hand must keep a $50$\,mm, $69.2$\,g cube in its grasp while imparting continuous rotational motion.
Physics runs at $100$\,Hz and the policy at $20$\,Hz; episodes last $500$ control steps, or $25$\,s, and terminate early if the vertical world coordinate of the cube centre falls below $-0.05$\,m.

\subsection{Observation and action}
The policy observation is restricted to signals available on the hardware.
At control step $t$ it is
\begin{equation}
  \mathbf{o}_t = \bigl(\, \tilde{\boldsymbol{\ell}}_t,\; \mathbf{a}_{t-1} \,\bigr) \in \mathbb{R}^{14},
  \qquad
  \tilde{\ell}_{t,i} = \ell_{t,i} + \varepsilon_{t,i},
  \label{eq:obs}
\end{equation}
where $\boldsymbol{\ell}_t \in \mathbb{R}^{7}$ are the measured channel values of Sec.~\ref{sec:notation}, $\mathbf{a}_{t-1}$ is the action applied at the previous step, and $\varepsilon_{t,i}$ is independent additive sensor noise drawn afresh at every step from $\mathcal{U}(-\bar\varepsilon_i, \bar\varepsilon_i)$, where $\mathcal{U}(a,b)$ denotes the uniform distribution on $[a,b]$, with $\bar\varepsilon_i = 5$\,mm on the six cable channels and $\bar\varepsilon_5 = 0.05$\,rad on the abduction channel.
Following the asymmetric actor--critic paradigm \cite{pinto2018asymmetric}, the value network additionally receives an $81$-dimensional privileged state.
It comprises the policy observation, the 16 joint angles and velocities, the seven actuator forces, the five fingertip positions relative to the palm, and the cube position error, orientation, angular velocity and linear velocity.
None of the privileged channels is perturbed.

The action $\mathbf{a}_t \in (-1,1)^{7}$ is the output of a hyperbolic tangent squashed Gaussian policy, and it sets the actuator targets relative to a fixed nominal grasp configuration $\boldsymbol{\ell}^{0} \in \mathbb{R}^{7}$,
\begin{equation}
  \boldsymbol{\ell}_t^{\mathrm{target}} = \boldsymbol{\ell}^{0} + \mathbf{s} \odot \mathbf{a}_t,
  \label{eq:action}
\end{equation}
where $\odot$ denotes the elementwise product, $\boldsymbol{\ell}^{0} = (0.09, 0.09, 0.09, 0.09, 0.75, 0.035, 0.1)$ is the keyframe used at every reset, and $\mathbf{s} = (0.02, 0.02, 0.02, 0.02, 0.7, 0.003, 0.012)$ collects the per-channel action scales.
Both vectors are expressed in metres on the six cable channels and radians on the abduction channel, following Sec.~\ref{sec:notation}.
The target $\boldsymbol{\ell}_t^{\mathrm{target}}$ is held for the whole control period and clamped by the simulator to the actuator range of each channel.
The scales are chosen so that $\boldsymbol{\ell}^{0} \pm \mathbf{s}$ stays inside every calibrated actuation range, which keeps the policy off the actuator limits; the thumb channels, whose ranges are three to five times narrower than a finger's, need correspondingly smaller scales.

\subsection{Reward}
The reward at each time step combines a rotation term and a fall penalty,
\begin{equation}
  r_t = \bigl( w_{\omega}\, \boldsymbol{\omega}_t^{\mathrm{cube}} \!\cdot\! \hat{\mathbf{z}}
      + w_{\mathrm{fall}}\, \mathbf{1}_{\mathrm{fall}}
      + w_{a}\, \lVert \mathbf{a}_t - \mathbf{a}_{t-1} \rVert_2^2 \bigr)\, \Delta t,
  \label{eq:reward}
\end{equation}
Here $\boldsymbol{\omega}_t^{\mathrm{cube}} \in \mathbb{R}^{3}$ is the angular velocity of the cube in the world frame in rad/s, $\hat{\mathbf{z}}$ is the world vertical unit vector, and $\Delta t = 0.05$\,s is the control period.
The indicator $\mathbf{1}_{\mathrm{fall}} \in \{0,1\}$ equals one on the step at which the vertical world coordinate of the cube centre falls below $-0.05$\,m, which also terminates the episode.
The weights are $w_{\omega} = 1$\,s/rad, $w_{\mathrm{fall}} = -100$ and $w_{a} = -1$.
The first term therefore rewards fast rotation about the vertical axis and the second penalises losing the grasp.
The third term penalises the rate of change of the command rather than a joint velocity.
The action sets actuator targets directly, so $\lVert \mathbf{a}_t - \mathbf{a}_{t-1}\rVert$ is proportional to the cable travel that the motors must produce within one $50$\,ms control period.
Bounding it is therefore the natural way to respect the maximum speed of the servos, and it also keeps the commands smooth enough to transfer.
The remaining terms available in the environment (cube linear velocity, pose regularisation, torque and energy costs) are weighted zero throughout.

\subsection{Domain randomisation}
To improve transfer we randomise the physical parameters listed in Table~\ref{tbl:dr}, resampled independently for every parallel environment \cite{tobin2017domain}.
The randomisation routine also draws a cube friction and a cube mass scale, but both are overwritten before the model is updated.
The reported policies were therefore trained with a fixed cube friction of $0.3$ and a fixed mass of $69.2$\,g.
We list them as not randomised rather than claim a robustness we did not train for.
The initial state of each episode is randomised as well.
The hand pose is perturbed with zero-mean Gaussian noise of standard deviation $0.1$\,rad, clipped to the joint limits.
The cube position is perturbed by $\pm 1$\,cm, and the cube orientation is drawn uniformly over $SO(3)$.

\begin{table}[!htbp]
    \centering
    \begin{tabular}{ll}
    \toprule
    Parameter & Distribution \\
    \midrule
    Fingertip friction (5 geoms)     & $\mathcal{U}(0.5,\,1.0)$ \\
    Cube inertia                     & $\times\,\mathcal{U}(0.8,\,1.2)$ \\
    Cube centre-of-mass offset       & $+\,\mathcal{U}(-5,\,5)$\,mm \\
    Joint zero offsets (16)          & $+\,\mathcal{U}(-0.05,\,0.05)$\,rad \\
    Joint dry friction (16)          & $\times\,\mathcal{U}(0.5,\,2.0)$ \\
    Joint armature (16)              & $\times\,\mathcal{U}(1.0,\,1.05)$ \\
    Link masses (21 bodies)          & $\times\,\mathcal{U}(0.9,\,1.1)$ \\
    Actuator position gain (7)       & $\times\,\mathcal{U}(0.8,\,1.2)$ \\
    Joint damping (16)               & $\times\,\mathcal{U}(0.8,\,1.2)$ \\
    \midrule
    Cube friction, cube mass         & not randomised (see text) \\
    \bottomrule
    \end{tabular}
    \caption{Domain randomisation actually applied during training.}
    \label{tbl:dr}
\end{table}

\begin{widetable}[!t]
    \centering
    \small
    \begin{tabular}{lcc}
    \toprule
    Setting & Training & Deployment \\
    \midrule
    Control rate                & $20$\,Hz & $20$\,Hz \\
    Cable action scale $s_{1:4}$& $0.020$\,m & $0.026$--$0.024$\,m \\
    Thumb scales $s_{5:7}$      & $0.7$, $0.003$, $0.012$ & identical \\
    Nominal $\ell^{0}_{1:4}$    & $0.090$\,m & $0.083$--$0.086$\,m \\
    Nominal $\ell^{0}_{5:7}$    & $0.75$, $0.035$, $0.1$ & identical \\
    Observation noise           & $\pm5$\,mm, $\pm0.05$\,rad & none \\
    \bottomrule
    \end{tabular}
    \caption{The learned controller is transferred unchanged; these are the only differences between the training environment and the Robot Operating System (ROS~2) node that runs on the hand. The slightly larger cable scales and the slightly more closed nominal pose were retuned on the physical hand.}
    \label{tbl:policycfg}
\end{widetable}

\subsection{Training}
We train for $3 \times 10^{8}$ environment steps with $8192$ parallel environments.
The policy and value networks are multilayer perceptrons with hidden layers $(512, 256, 128)$ and swish activations.
We use a learning rate of $3 \times 10^{-4}$, a discount factor of $0.97$ and an entropy cost of $10^{-2}$.
Each batch is collected with an unroll length of $40$, then split into $32$ minibatches of size $256$ and used for $4$ update epochs.
Observation normalisation is enabled.
The remaining PPO parameters keep the Brax defaults, namely a clipping parameter of $0.3$, a generalised advantage estimation factor $\lambda = 0.95$, and no gradient clipping.
\begin{reportonly}
Fig.~\ref{fig:rewardcurve} shows the training curve.

\begin{figure}[!htbp]
    \centering
    \includegraphics[width=0.98\linewidth]{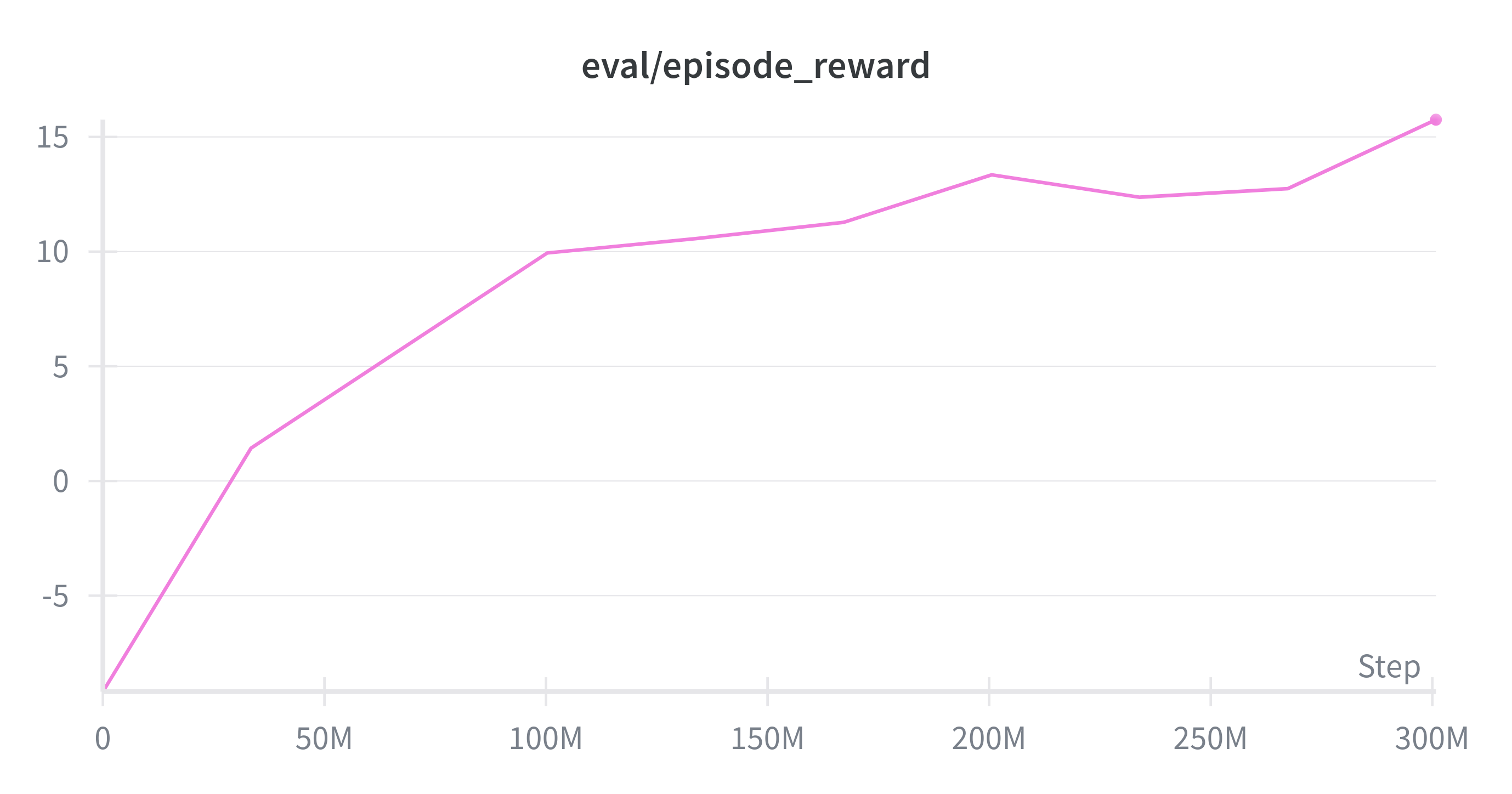}
    \caption{Episode reward during training of the $z$-axis cube rotation task, over $3 \times 10^{8}$ environment steps with $8192$ parallel environments. Each marker is one evaluation.}
    \label{fig:rewardcurve}
\end{figure}
\end{reportonly}

\section{Results}
\label{sec:results}

\subsection{Behaviour in simulation}
\begin{figure}[!htbp]
    \centering
    \ifdefined\icraversion
      \includegraphics[width=0.98\linewidth]{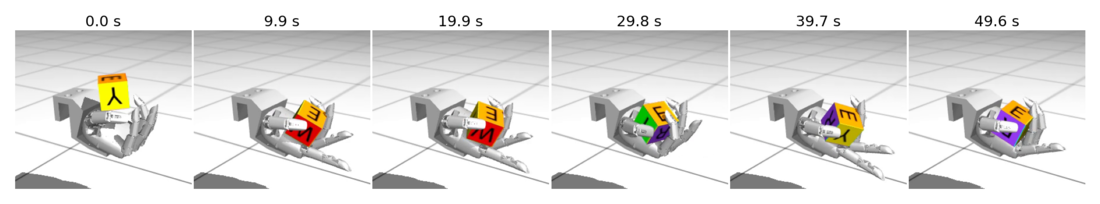}
    \else
      \includegraphics[width=0.98\linewidth]{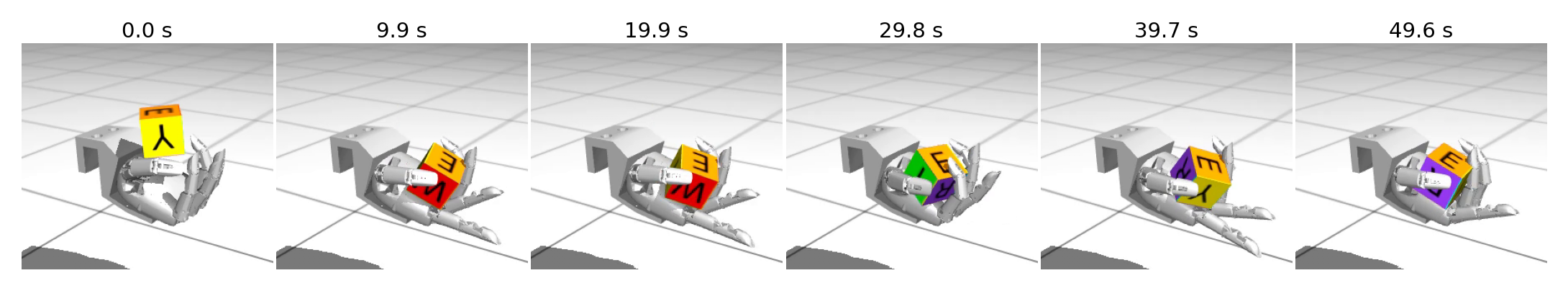}
    \fi
    \caption{The learned policy rotating the cube in simulation, in frames from a continuous $50$\,s rollout; the quantitative evaluation in the text is a separate $15$\,s episode. The cube faces track the rotation about the vertical axis; the grasp is maintained throughout.}
    \label{fig:simrotation}
\end{figure}

The policy rotates the cube continuously in simulation (Fig.~\ref{fig:simrotation}).
For a quantitative measure we evaluate it in the plain-MuJoCo replica of the environment, under the deployment settings of Table~\ref{tbl:policycfg} and started from the nominal grasp.
It holds the cube for the full $15$\,s evaluation episode and turns it through $7.64$\,rad.
That is $1.22$ revolutions, at a mean rate of $0.51$\,rad/s.

\subsection{Deployment without policy fine-tuning}
\label{sec:deploy}
\begin{figure}[!htbp]
    \centering
    \ifdefined\icraversion
      \includegraphics[width=0.98\linewidth]{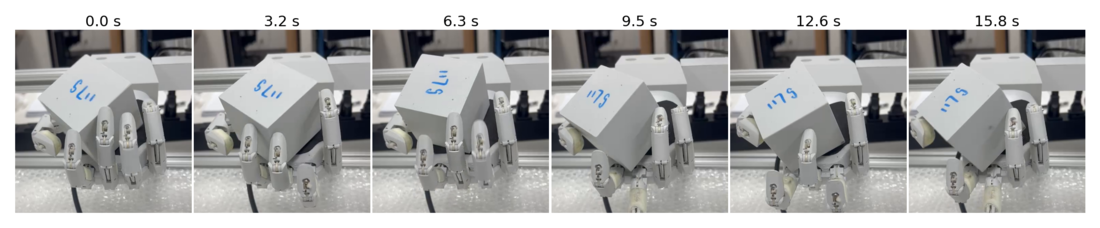}
    \else
      \includegraphics[width=0.98\linewidth]{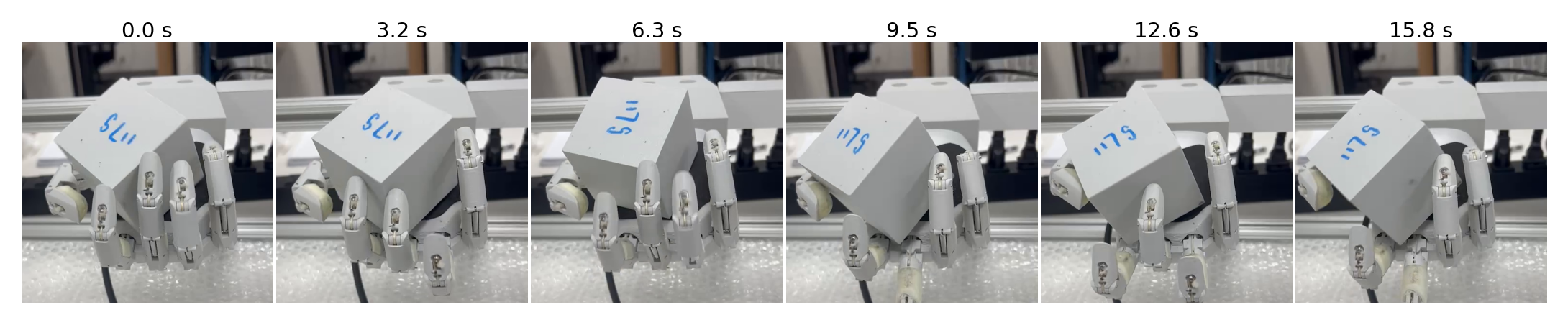}
    \fi
    \caption{Deployment on the physical hand. The hand rotates the cube about the $z$-axis over a $16$\,s recording, with no policy fine-tuning and no sensing beyond the seven motor encoders. The marking on the cube face tracks the rotation.}
    \label{fig:realrotation}
\end{figure}

The policy is transferred to the physical hand without any fine-tuning, as a ROS~2 node running at the same $20$\,Hz as training.
At each control step the node reads the seven actuator states and maps them into simulation units, the four finger cables and the abduction joint through the affine map \eqref{eq:servomap} and the two thumb cables by solving \eqref{eq:thumbinv}, and assembles the same $14$-dimensional observation used during training.
The deterministic policy output is converted to cable-length targets through \eqref{eq:action} and mapped back to actuator commands, the fingers through the inverse of \eqref{eq:servomap} and the thumb through \eqref{eq:thumbfit} followed by the joint-to-actuation model of \eqref{eq:thumbcmc} and \eqref{eq:thumbflexor}.
Because the policy was trained exclusively on hardware-available signals, no state estimation and no additional sensing enter the loop.
The policy transfers without fine-tuning and sustains continuous cube rotation on the physical hand (Fig.~\ref{fig:realrotation}).
\rtwo{The cube pose is not logged on the hardware, so the rate is not measured there; in the recording the cube turns at roughly one revolution per $55$\,s, and the grasp is held throughout.}

Table~\ref{tbl:policycfg} lists the only differences between the training environment and the deployed node.
\rtwo{Both run at $20$\,Hz, and the thumb channels share the same scales and nominal values.}
During commissioning on the physical hand the cable scales were opened up by $20$--$30\%$ and the nominal pose was closed by $4$--$8$\,mm.
These are constants of the deployment node's action map rather than of the policy, whose weights are unchanged; the transfer is therefore free of policy fine-tuning, not of hardware-side calibration.
The observation noise used as a training augmentation is naturally absent at deployment.

\subsection{Separating transmission mismatch from policy feedback}
To attribute the residual gap, we additionally logged the action sequence of a simulation rollout and replayed it open loop on both the simulator and the physical hand.
With the policy held out of the loop, agreement between the two replays indicates that the residual sim-to-real gap originates from the policy feedback rather than from the transmission model, and disagreement indicates the converse.
Combined with the per-channel residuals of Table~\ref{tbl:validation}, this localises the remaining error to the thumb.
The four finger channels replay to within a millimetre, while the thumb channels carry both the largest excursion mismatch and the largest tracking residual.

\section{Conclusion and Outlook} 
\label{sec:conclusion}
We presented \handname, a tendon-driven dexterous hand that \rtwo{performs all $33$ grasp types of the GRASP taxonomy} at $374$\,g and that is fully 3D printed and open-sourced.
The contribution of this work is the pipeline that makes such a hand, whose cable transmission is invisible to a conventional joint-actuated simulation model, learnable in simulation.
It consists of a simulation model in which the transmission itself is represented, an identified actuation map that connects the model to the motor commands in both directions, and a training package whose observation and action spaces are restricted to what the hardware provides.

We model the routing geometry instead of fitting moment arms.
The kinematic agreement we report, $0.04$--$1.40$\,mm of cable excursion on the four fingers, is therefore a prediction of the model rather than a fitted quantity.
The dynamic residual of $0.29$--$0.45$\,mm RMS under a shared command stream shows that the agreement survives motion.
Restricting the learned policy to the seven signals the hardware actually provides then makes deployment a matter of applying the identified map, with no state estimation in the loop.
An in-hand cube rotation policy trained entirely in this model transfers without fine-tuning.

We intend to extend the released task suite beyond in-hand rotation, so that the model is exercised by \rtwo{grasps drawn from the full range the hand can reach}.
The hardware, the hand model, the simulation environment and the analysis scripts that produce every number reported here are open-sourced.

\begin{reportonly}
\Needspace*{12\baselineskip}
\section{\rtwo{Resources}}
\label{sec:resources}
\begin{rtwoblock}
\grouphead{Project website}
\url{https://chestnut-robotics.github.io/aero-hand-open/}

\grouphead{Open-source repository}
\url{https://github.com/chestnut-robotics/aero-hand-open}

\grouphead{Online documentation}
\url{https://docs.tetheria.ai/}
\end{rtwoblock}
\end{reportonly}

\ifdefined\icraversion
  \bibliographystyle{IEEEtran}
\else
  \bibliographystyle{unsrtnat}
\fi
\bibliography{references}

\end{document}